\documentclass[3p]{elsarticle}

\usepackage{times}
\usepackage{latexsym}
\usepackage{booktabs}
\usepackage{graphicx}
\usepackage{multirow}
\usepackage{times}
\usepackage{amsfonts,amsmath,amssymb,bm}
\usepackage{graphicx}
\usepackage{xcolor}
\usepackage{setspace}
\usepackage{pdflscape}
\usepackage{pdfpages}
\usepackage{enumitem}
\usepackage{rotating}
\usepackage{caption}
\usepackage{mathabx}
\usepackage{subcaption}
\usepackage{hyperref}
\usepackage{caption}
\usepackage{subcaption}

\usepackage[english]{babel}

\addto\captionsenglish{%
}
\addto\extrasenglish{%
}

\journal{Artificial Intelligence}
\biboptions{sort&compress}
\begin{document}

\begin{frontmatter}



\title{RelBERT: Embedding Relations with Language Models}


\author{Asahi Ushio}
\author{Jose Camacho-Collados}
\author{Steven Schockaert}

\affiliation{organization={Cardiff NLP, School of Computer Science and Informatics, Cardiff University},
            addressline={Senghennydd Rd}, 
            city={Cardiff},
            postcode={CF24 4AG}, 
            country={United Kingdom}}

%

\begin{abstract}
Many applications need access to background knowledge about how different concepts and entities are related. Although Knowledge Graphs (KG) and Large Language Models (LLM) can address this need to some extent, KGs are inevitably incomplete and their relational schema is often too coarse-grained, while LLMs are inefficient and difficult to control. As an alternative, we propose to extract relation embeddings from relatively small language models. In particular, we show that masked language models such as RoBERTa can be straightforwardly fine-tuned for this purpose, using only a small amount of training data. The resulting model, which we call RelBERT, captures relational similarity in a surprisingly fine-grained way, allowing us to set a new state-of-the-art in analogy benchmarks. Crucially, RelBERT is capable of modelling relations that go well beyond what the model has seen during training. For instance, we obtained strong results on relations between named entities with a model that was only trained on lexical relations between concepts, and we observed that RelBERT can recognise morphological analogies despite not being trained on such examples. Overall, we find that RelBERT significantly outperforms strategies based on prompting language models that are several orders of magnitude larger, including recent GPT-based models and open source models. \footnote{Source code to reproduce our experimental results and the model checkpoints are available in the following repository:  \url{https://github.com/asahi417/relbert}.}

\end{abstract}




\end{frontmatter}


\section{Introduction}\label{sec:introduction}
Recognizing the lexical relationship between two words has long been studied as a fundamental task in natural language processing (NLP) \cite{Turney:2005:MSS:1642293.1642475}.  As a representative early example, DIRT \cite{Lin2001} first collects sentences in which two given target words co-occur (e.g.\ \emph{London} and \emph{U.K.}) and then uses the dependency paths between the two words to model their relationship. Along similar lines, {Latent Relational Analysis (LRA \cite{Turney:2005:MSS:1642293.1642475}) relies on templates expressing lexical patterns to characterise word pairs (e.g.\ \emph{[head word] is the capital of [tail word]}), thus again relying on sentences where the words co-occur. After the advent of word embeddings \cite{mikolov2013distributed,pennington-etal-2014-glove,bojanowski2016enriching}, most approaches for modelling relations relied on word vectors in one way or another. A common strategy to model the relation between two words was to take the vector difference between the embeddings of each word \cite{mikolov2013distributed,gladkova-etal-2016-analogy,vylomova-etal-2016-take}. For example, the relationship between ``King'' and ``Queen'' is the gender difference, which can be captured by $\mathsf{wv}({\rm King}) - \mathsf{wv}({\rm Queen})$, where $\mathsf{wv}(X)$ denotes the embedding of word $X$. Although the vector difference of word embeddings quickly gained popularity, it has been shown that the latent space of such relation vectors is noisy, with nearest neighbours often corresponding to different relationships \cite{linzen-2016-issues,drozd-etal-2016-word,bouraoui-etal-2018-relation}. The limitations of word embedding differences can also clearly be seen on SAT \cite{DBLP:conf/ranlp/TurneyLBS03}, a well-known benchmark involving word pair analogies (see \autoref{analogy-questions}), where the accuracy of word vector differences is particularly disappointing \cite{ushio-etal-2021-bert}.

Knowledge graphs (KGs) such as Wikidata \cite{vrandevcic2014wikidata} and ConceptNet \cite{conceptnet2017} are also closely related to the study of relation understanding. In contrast to the aforementioned methods, KGs rely on symbolic representations. They use a fixed relational schema to explicitly encode the relationships between words or entities. KGs are more interpretable than embeddings, but they usually have the drawback of being highly incomplete.
Moreover, due to their use of a fixed relational schema, KGs are inherently limited in their ability to capture subtle and fine-grained differences between relations, which is essential for many knowledge-intensive tasks. For example, \autoref{tab:relbert:sat-sample} shows two instances of 
 the SAT analogy task, where the relationships found in the query are abstract. When trying to solve such questions with a KG, we may have access to triples such as (\emph{wing}, \emph{UsedFor}, \emph{air}), but this kind of knowledge is not sufficient to solve the given question, e.g.\ since all of (\emph{lung}, \emph{UsedFor}, \emph{breath}), (\emph{engine}, \emph{UsedFor}, \emph{jet}), and (\emph{flipper}, \emph{UsedFor}, \emph{water}) make sense. The issue here is that the relationship \emph{UsedFor} is too vague and does not describe the relationship between \emph{wing} and \emph{air} accurately enough to solve the task. Such limitations of KGs have recently been tackled with pre-trained language models (LMs) \cite{devlin-etal-2019-bert,GPT3,2020t5}, which have been shown to implicitly capture various forms of factual knowledge  \cite{petroni-etal-2019-language,jiang-etal-2020-know,heinzerling-inui-2021-language}. In particular, some researchers have proposed to extract KG triples from LMs \cite{west-etal-2022-symbolic,hao2022bertnet,cohen2023crawling}, which offers a compelling strategy for automatically enriching existing KGs. 
However, the resulting representations are still too coarse-grained for many applications. LLMs are also inefficient and difficult to control. 

\begin{table}[t]
\centering
\begin{tabular}[t]{lcl}
\toprule
Query:       && wing:air \\
\midrule
Candidates: & (1) & arm:hand \\
 & (2) & lung:breath \\
 & \textbf{(3)} & \textbf{flipper:water} \\
 & (4) & cloud:sky \\
 & (5) & engine:jet\\
\bottomrule
\end{tabular}
\hspace{1em}
\begin{tabular}[t]{lcl}
\toprule
Query:       && perceptive:discern \\
\midrule
Candidates: & (1) & determined:hesitate \\
 & (2) & authoritarian:heed \\
 & (3) & abandoned:neglect \\
 & (4) & restrained:rebel \\
 & \textbf{(5)} & \textbf{persistent:persevere} \\
\bottomrule
\end{tabular}
\caption{Two examples of analogy task from the SAT dataset, where the candidate in bold characters is the answer in each case.}
\label{tab:relbert:sat-sample}
\end{table}

In this paper, we propose a framework to distill relational knowledge from a pre-trained LM in the form of relation embeddings. 
Specifically, we obtain the relation embedding of a given word pair by feeding that word pair to the LM using a fixed template and by aggregating the corresponding contextualised embeddings.
We fine-tune the LM using a contrastive loss, in such a way that the relation embeddings of word pairs that have a similar relationship become closer together, while moving further away from the embeddings of unrelated word pairs. Our main models, which we refer to as \emph{RelBERT}, are based on RoBERTa \cite{RoBERTa} as the underlying LM and are fine-tuned on a modified version of a dataset about relational similarity from SemEval 2012 Task 2 \cite{jurgens-etal-2012-semeval}. 
Despite the conceptual simplicity of this approach, the resulting model outperforms all the baselines including LRA \cite{Turney:2005:MSS:1642293.1642475}, word embeddings \cite{mikolov2013distributed,pennington-etal-2014-glove,bojanowski2016enriching}, and large scale language models such as OPT \cite{zhang2022opt,iyer2022opt} and T5 \cite{2020t5,https://doi.org/10.48550/arxiv.2210.11416} in the zero-shot setting (i.e.\ without any task-specific validation or additional model training). For example, RelBERT achieves 73\% accuracy on the SAT analogy benchmark, which outperforms the previous state-of-the-art by 17 percentage points, and GPT-3 \cite{GPT3} by 20 percentage points (in the zero-shot setting). 
The strong performance of RelBERT is especially remarkable given the small size of the considered training set.
Moreover, being based on relatively small language models, RelBERT is surprisingly efficient. In fact, the version of RelBERT based on RoBERTa\textsubscript{BASE} has 140 million parameters, but already outperforms GPT-3, which has 175 billion parameters, across all the considered benchmarks. Overall, we test RelBERT in nine diverse analogy benchmarks and find that it achieves the best results in all cases. Crucially, the fixed-length vector formulation of RelBERT enables a flexibility not found in standard language models, and can be leveraged for multiple applications. For instance, RelBERT proves competitive to the state-of-the-art in lexical relation classification, outperforming all previous embedding approaches.

To further understand the capability of RelBERT, we analyse its performance from various perspectives. One important finding is that RelBERT performs strongly even on relation types that are fundamentally different from the ones in the training data. For instance, while the training data involves standard lexical relations such as hypernymy, synonymy, meronyny and antonymy, the model is able to capture morphological relations, and even factual relations between named entities. Interestingly, our standard RelBERT model, which is trained on the relation similarity dataset, achieves similar or better results for the latter type of relations than models that are trained on relations between named entities. Moreover, even if we remove all training examples for a given lexical relation (e.g.\ hypernymy), we find that the resulting model is still capable of modelling that relationship. These findings highlight the generalisation ability of RelBERT for recognizing word pairs with unseen relation types by extracting relation knowledge from the pre-trained LM, rather than merely generalising the examples from the training data.

\paragraph{Motivation}
Relations play a central role in many applications. For instance, many question answering models currently rely on ConceptNet for modelling the relation between the concepts that are mentioned in the question and a given candidate answer \cite{yasunaga-etal-2021-qa,sun-etal-2022-jointlk,jiang-etal-2022-great}. Commonsense KGs are similarly used to provide additional context to computer vision systems, e.g.\ for generating scene graphs \cite{DBLP:conf/cvpr/GuZL0CL19,DBLP:conf/wacv/ChenRL23} and for visual question answering \cite{DBLP:conf/aaai/WuLSM22}. Many recommendation systems also rely on knowledge graphs to identify and explain relevant items \cite{DBLP:conf/www/WangZXG18,DBLP:conf/kdd/Wang00LC19}. Other applications that rely on knowledge graphs, or on modelling relationships more broadly, include semantic search \cite{info:doi/10.2196/16948,DBLP:conf/sigir/JafarzadehAE22}, flexible querying of relational databases \cite{bordawekar2017using}, schema matching \cite{fernandez2018seeping}, completion and retrieval of Web tables \cite{zhang2019table2vec} and ontology completion \cite{DBLP:conf/aaai/BouraouiS19}. Many of the aforementioned applications rely on knowledge graphs, which are incomplete and limited in expressiveness due to their use of a fixed relation schema. Relation embeddings have the potential to address these limitations, especially in contexts which involve ranking or measuring similarity, where extracting knowledge by prompting large language models (LLMs) cannot replace vector-based representations. Relation embeddings can also provide a foundation for systems that rely on analogical reasoning, where we need to identify correspondences between a given scenario and previously encountered ones \cite{gentner1997structure}. Finally, by extracting relation embeddings from LMs, we can get more insight into what knowledge is captured by such models, since these embeddings capture the knowledge from the model in a more direct way than what is possible with prompting based methods \cite{petroni-etal-2019-language,jiang-etal-2020-know}. Indeed, the common prediction-based model probing techniques \cite{petroni-etal-2019-language,jiang-etal-2020-know} can easily be manipulated by adversarial inputs, e.g.\ involving negation \cite{kassner-schutze-2020-negated}. Accordingly, recent studies have focused on identifying language model parameters that represent factual knowledge about named entities \cite{meng2022locating}. 
We believe that relation embeddings can be seen in a similar light, offering the potential for more direct analysis of the knowledge captured by language models.

\paragraph{Structure of the paper}
After discussing the related work in \autoref{secRelatedWork}, we first introduce RelBERT, our framework for extracting relation embeddings from fine-tuned LMs, in \autoref{sec:relbert:relbert}. We then describe our two main evaluation tasks in \autoref{sec:relbert:evaluation-tasks}: analogy questions and lexical relation classification. \autoref{experimental-setting} presents our experimental setup. Subsequently, \autoref{sec:relbert:experimental-results} compares the results of RelBERT with baselines, including a large number of recent LLMs. To better understand the generalisation ability of RelBERT, in \autoref{sec:relbert:training-data-overlap} we conduct an experiment in which certain relation types are excluded from the training set and then evaluate the model on the excluded relation type. In addition to the main experiment, we compare RelBERT with conversational LMs and few-shot prompting strategies in \autoref{sec:relbert:additional-baselines}. As the learning process of RelBERT can be affected by many factors, we provide a comprehensive analysis of RelBERT fine-tuning in \autoref{sec:relbert:ablation-abalysis}. Finally, we provide a qualitative analysis of the relation embedding space of RelBERT \autoref{sec:relbert:qualitative-analysis}. 

This paper extends our earlier conference paper \cite{ushio-etal-2021-distilling} in several ways: 1) we now consider two additional losses for training RelBERT; 2) we evaluate on four additional benchmarks; 3) we consider several alternative training sets; 4) we extensively compare against recent language models of sizes up to 30B parameters; 5) we analyse the role of training data in the RELBERT fine-tuning process. We find that, surprisingly, RelBERT achieves a non-trivial performance on named entities, despite only being trained on concepts. Moreover, on analogies between concepts, even the smallest RelBERT model, with 140M parameters, substantially outperforms all the considered LMs.


\section{Related Work}\label{secRelatedWork}

\subsection{Unsupervised Relation Discovery}
Modelling how different words are related is a long-standing challenge in NLP. An early approach is DIRT \cite{Lin2001}, which encodes the relation between two nouns as the dependency path connecting them. The idea is that two such dependency paths are similar if the sets of word pairs with which they co-occur are similar. Along the same lines, \cite{DBLP:conf/acl/HasegawaSG04} cluster named entity pairs based on the bag-of-words representations of the contexts in which they appear. In \cite{Yao2011}, a generative probabilistic model inspired by LDA \cite{DBLP:journals/jmlr/BleiNJ03} was proposed, in which relations are viewed as latent variables (similar to topics in LDA). Turney \cite{Turney:2005:MSS:1642293.1642475} proposed a method called latent relational analysis (LRA), which uses matrix factorization to learn relation embeddings based on co-occurrences of word pairs and dependency paths. Matrix factorization is also used in the Universal Schema approach from Riedel et al. \cite{Riedel2013}, which represents entity pairs by jointly modelling (i) the contexts of occurrences of entity pairs in a corpus and (ii) the relational facts that are asserted about these entities in a given knowledge base. After the introduction of Word2Vec, several approaches were proposed that relied on word embeddings for summarising the contexts in which two words co-occur. For instance, \cite{Jameel2018} introduced a variant of the GloVe word embedding model, in which relation vectors are jointly learned with word vectors. In SeVeN \cite{espinosa-anke-schockaert-2018-seven} and RELATIVE \cite{camachocollados:ijcai2019relative}, relation vectors are computed by averaging the embeddings of context words, while pair2vec \cite{joshi-etal-2019-pair2vec} uses an LSTM to summarise the contexts in which two given words occur, and \cite{Washio2018a} learns embeddings of dependency paths to encode word pairs. Another line of work is based on the idea that relation embeddings should facilitate link prediction, i.e.\ given the first word and a relation vector, we should be able to predict the second word \cite{Marcheggiani2016,DBLP:conf/acl/SimonGP19}.

\subsection{Language Models for Relational Knowledge}
The idea of extracting relational knowledge from pre-trained LMs has been extensively studied. For instance, \cite{petroni-etal-2019-language} uses BERT for link prediction. They use a manually defined prompt for each relation type, in which the tail entity is replaced by a \texttt{<mask>} token. To complete a knowledge graph triple such as (\emph{Dante}, \emph{born-in}, ?) they create the input ``\emph{Dante was born in \texttt{<mask>}}'' and then look at the predictions of BERT for the masked token to retrieve the correct answer. The results of this analysis suggest that BERT captures a substantial amount of factual knowledge, a finding which has inspired a line of work in which LMs are viewed as knowledge bases. Later, the analysis from \cite{petroni-etal-2019-language} has been improved by adding instances with negation in \cite{kassner-schutze-2020-negated}, and extended to non-English languages in \cite{kassner-etal-2021-multilingual}. Some works have also looked at how relational knowledge is stored. In \cite{geva2020transformer}, it is argued that the feed-forward layers of transformer-based LMs act as neural memories, which would suggest that e.g.\ ``the place where Dante is born'' is stored as a property of Florence. Some further evidence for this view is presented in \cite{dai2021knowledge}. What is less clear is whether relations themselves have an explicit representation, or whether transformer models essentially store a propositionalised knowledge graph. The results we present in this paper suggest that common lexical relations (e.g.\ hypernymy, meronymy, has-attribute), at least, must have some kind of explicit representation, although it remains unclear how they are encoded. In \cite{bouraoui2020inducing}, they analyse the ability of BERT to identify word pairs that belong to a given relation. In our earlier work \cite{ushio-etal-2021-bert}, we have evaluated the ability of LMs to directly solve analogy questions. The main finding was that LMs are poor at solving analogy questions with a vanilla perplexity based approach, although results can be improved with a carefully-tuned scoring function. In \cite{rezaee-camacho-collados-2022-probing}, they extended this analysis by evaluating the sensitivity of language models to the direction of a word pair (e.g.\ by checking whether the model can distinguish the word pair \emph{London}:\emph{U.K.} from the word pair \emph{U.K.}:\emph{London}),  the ability to recognize which entity type can form a specific relation type (e.g.\ the head and tail entity of the \emph{born-in} relation should be person and location) and the robustness to some adversarial examples. Their main findings were that LMs are capable of understanding the direction and the type of a relationship, but can be distracted by simple adversarial examples. For instance, both \emph{Paris}:\emph{France} and \emph{Rome}:\emph{France} were predicted to be instances of the \emph{capital-of} relation.

Given the observation that LMs capture an extensive amount of relational knowledge, LMs have been used for tasks such as KG completion, and even for generating KGs from scratch. For instance, in \cite{davison-etal-2019-commonsense}, a triple is first converted into a sentence, by choosing a template based on the log-likelihood estimates of a causal language model (CLMs). The resulting sentence is then fed into a masked LM to estimate the plausibility of the triple based on the log-likelihood of the masked token prediction of the head and tail words. However, this approach is inefficient to use in practice, since all the candidate triples have to be tested one by one. 
To avoid such issues, \cite{wang2020language} proposed to directly extract plausible triples using a pre-trained LM. Given a large corpus such as Wikipedia, they parse every sentence in the corpus to find plausible triples with a pre-trained LM. First, a single sentence is fed to an LM to obtain the attention matrix, and then for every combination of two words in the sentence, they find intermediate tokens in between the two words, which contribute to predict the two words, by decoding the attention matrix. In the end, the word pairs are simply filtered by the corresponding attention score, and the resulting word pairs become the triples extracted from the sentence, where the intermediate tokens of each pair are regarded as describing the relationship. 
Instead of extracting triples from a corpus, \cite{alivanistos2022prompting} proposed to use LMs to complete a triple by generating a tail word given a head word and a relation. They manually create a number of templates for each relation type, where a single template contains a placeholder to be filled by a head word. Each template is fed to a pre-trained LM to predict the tail word. As a form of post-filtering, they use a pre-trained LM to score the factual validity of the generated triples with another prompt to enhance the precision. Unlike the method proposed in \cite{wang2020language}, which extracts an entire triple, \cite{alivanistos2022prompting} assumes that the head and the relation are given, so it is more suited to KG completion, while \cite{wang2020language} is rather aimed at constructing KGs from scratch.
Recently, \cite{huang-etal-2022-deer} proposed a two-step process for learning a KG in which relations are represented as text descriptions. In the first step, sentences in Wikipedia that explicitly describe relations between entities are identified. To improve the coverage of the resource, in the second step, T5 \cite{2020t5} is used to introduce additional links. Specifically, they use a fusion-in-decoder \cite{izacard-grave-2021-leveraging} to generate descriptions of the relationship between two entities, essentially by summarising the descriptions of the paths in the original KG that connect the two entities.

Where the aforementioned works extract KGs from LMs, conversely, there has also been a considerable amount of work on infusing the knowledge of existing KGs into LMs. Early approaches introduced auxiliary tasks that were used to train the LM alongside the standard language modelling task, such as entity annotation \cite{logan-etal-2019-baracks} and relation explanation \cite{hayashi2020latent} based on KGs. ERNIE \cite{sun2019ernie} is a masked LM similar to BERT, but they employ a masking strategy that focuses on entities that are taken from a KG, unlike BERT, which randomly masks tokens during pre-training. In addition to the entity-aware masking scheme, LUKE \cite{yamada-etal-2020-luke} conditions internal self-attention by entity-types. It achieved better results than vanilla LMs in many downstream tasks. Since it is computationally demanding to train LMs from scratch, there is another line of work that relies on fine-tuning existing LMs. For instance, \cite{peters-etal-2019-knowledge} fine-tuned BERT based on the cross-attention between the embeddings from BERT and an entity linking model. Their model learned a new projection layer to generate entity-aware contextualized embeddings. 

\subsection{Modelling Analogy}
Modelling analogies has a long tradition in the NLP community. The aforementioned LRA model \cite{Turney:2005:MSS:1642293.1642475}, for instance, was motivated by the idea of solving multiple-choice analogy questions. Despite its simplicity, LRA achieved a strong performance on the SAT benchmark, which even GPT-3 is not able to beat in the zero-shot setting \cite{GPT3}.
The idea of using word vector differences for identifying analogies was popularised by \cite{mikolov2013distributed}. The core motivation of using word embeddings for modelling analogies dates back to connectionism theory \cite{feldman1982connectionist}, where neural networks were thought to be capable of learning emergent concepts \cite{hopfield1982neural, hinton1986learning} with distributed representations across a semantic embedding space \cite{hinton1986distributed}. More recent works have proposed mathematical justifications and experiments to understand the analogical reasoning capabilities of word embeddings, by attempting to understand their linear algebraic structure \cite{arora2016latent,gittens2017skip,allen2019analogies} and by explicitly studying their compositional nature \cite{levy2014linguistic,paperno2016whole,ethayarajh2019towards,chiang-etal-2020-understanding}. 

Recently, the focus has shifted to modelling analogies using LMs. For instance, \cite{chen-etal-2022-e} proposed E-KAR, a benchmark for analogy modelling which essentially follows the same multiple-choice format as SAT, except that an explanation is provided for why the analogy holds and that some instances involve word triples rather than word pairs. In addition to the task of solving analogy questions, they also consider the task of generating explanations for analogies. Both tasks were found to be challenging for LMs. 
In \cite{bhavya-etal-2022-analogy}, they used prompt engineering to generate analogies with GPT-3. They consider two analogy generation tasks: (i) generating an explanation with analogies for a target concept such as ``Explain Bohr's atomic model using an analogy'', and (ii) generating an explanation of how two given concepts are analogous to each other such as ``Explain how Bohr's atomic model is analogous to the solar system''. They argue that GPT-3 is capable of both generating and explaining analogies, but only if an optimal prompt is chosen, where they found the performance to be highly sensitive to the choice of prompt.
In \cite{sultan-shahaf-2022-life}, they used LMs to find analogies between the concepts mentioned in two documents describing situations or processes from different domains. 
To improve the quality of analogies generated by LMs, \cite{DBLP:conf/www/BhavyaXZ23} proposed an LM based scoring function to detect low-quality analogies. They start from manually-crafted templates that contain the information of the domain (e.g. ``Machine Learning'') and the target concept (e.g. ``Language Model''). The templates are designed so that LMs can generate explanations of the target concept involving analogies. Once they generate analogies with the templates, they evaluate the generated analogies from the perspectives of analogical style, meaningfulness, and novelty, to identify which analogies to keep. The evaluation of the analogies is then used to improve the templates, and the low-quality analogies are re-generated with the improved templates. The evaluation relies on automatic metrics, and the template re-writing is done via prompting to edit the current template with the feedback, so the process can be iterated to repeatedly improve low-quality analogies.
In \cite{10.1145/3543873.3587333}, they fine-tuned masked LMs to solve analogy questions. 
The embedding for a given word pair $(w_1,w_2)$ is obtained as the contextualised representation of the \texttt{<mask>} token with the prompt ``$w_1$ \texttt{<mask>} $w_2$''. To fine-tune LMs on analogy questions, they convert the task into a binary classification of $A$:$B$::$C$:$D$ as an analogy or not, where ($A$,$B$) is the query word pair and ($C$,$D$) is a candidate word pair. With the binary analogy classification formulation, they fine-tune an LM with a linear layer on top of the word pair embeddings of query and candidate word pairs. They use the resulting fine-tuned model to annotate more instances as a form of data augmentation and continue to fine-tune the model on the generated pseudo dataset.


\section{RelBERT} \label{sec:relbert:relbert}
We now introduce our proposed RelBERT model, a fine-tuned LM encoder of the BERT family for modelling relational similarity. The input to RelBERT consists of a word pair, which is fed to the LM using a prompt. The LM itself is fine-tuned to map this input to a vector that encodes how the two given words are related. We will refer to this vector as a \emph{relation embedding}. A schematic overview of the RelBERT model is shown in \autoref{fig:relbert:image_embedding}. Our overall strategy is explained in more detail in \autoref{sec:relbert:overall-strategy}, while the details of the fine-tuning process are provided in \autoref{sec:relbert:relational-knowledge-distillation-via-language-model-finetuning}.

\begin{figure}[t]
    \centering
    \includegraphics[width=\columnwidth]{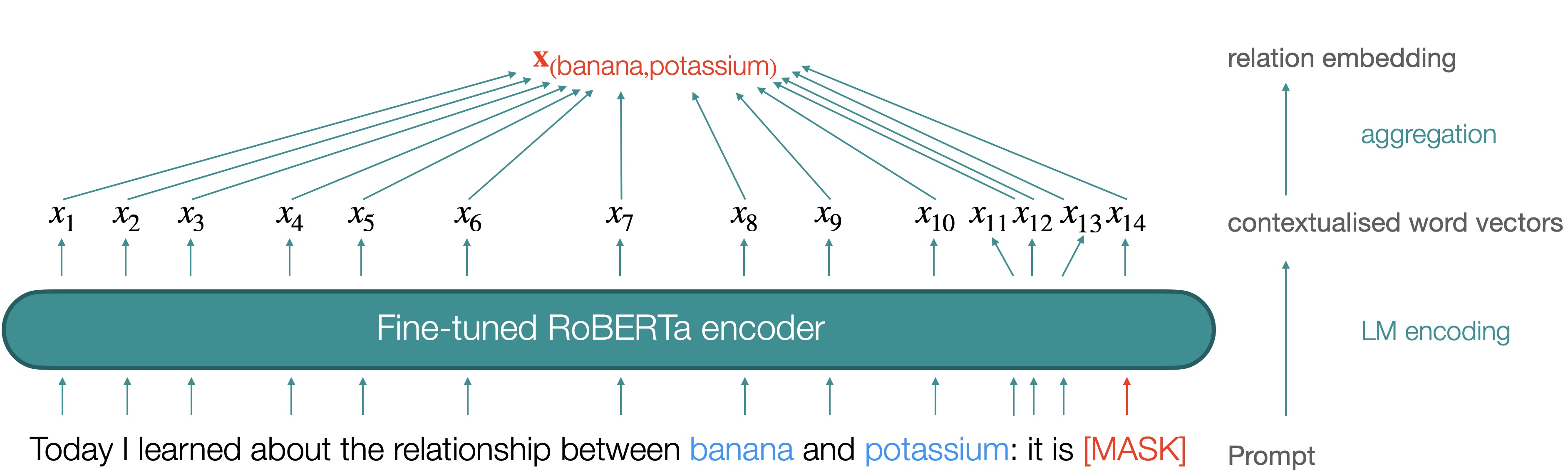}
    \caption{Schematic overview of the RelBERT model. A word pair is presented to an LM encoder using a prompt. A relation vector, capturing how the two input words are related, is then obtained by aggregating the contextualised embeddings from the output layer.}
    \label{fig:relbert:image_embedding}
\end{figure}

\subsection{Overall Strategy} \label{sec:relbert:overall-strategy}
To obtain the relation embedding of a word pair $(h, t)$, we need to construct a suitable input for the language model. While it is possible to simply use the pair $(h,t)$ as input, similar to what is done by COMET \cite{bosselut-etal-2019-comet}, better results can be achieved by converting the word pair into a more or less naturally sounding sentence. This is true, in particular, because the amount of high-quality data that is available for training RelBERT is relatively limited, as we will see in \autoref{sec:relbert:dataset}. We thus need to manually create a template with placeholders for the two target words, which somehow expresses that we are interested in modelling the relationship between the two words. Such a strategy has already been proven effective for factual knowledge probing \cite{petroni-etal-2019-language} and text classification \cite{schick-schutze-2021-exploiting,tam2021improving,le2021many}, among many others. Since we will rely on fine-tuning the LM, the exact formulation of the prompt matters less than in zero-shot settings. However, we found that performance suffers when the prompt is too short, in accordance with \cite{bouraoui2020inducing} and \cite{jiang-etal-2020-know}, or when the prompt is nonsensical (i.e.\ when it does not express the idea of modelling a relationship). With this in mind, we will use the following five templates for our main experiments\footnote{In our previous paper \cite{ushio-etal-2021-distilling}, we evaluated different types of prompts, including automatically-generated ones. For this paper, we tried to extend this initial analysis but the results were inconclusive. This suggests that the choice of prompt may be somewhat less important than we had initially assumed. This view is also supported by the recent analysis in in \cite{pitarch-etal-2023-clues}, which showed that the LM can be successfully fine-tuned to learn relation embeddings with short and uninformative prompts, with only a very small degradation in quality. We will come back to the analysis of prompt importance in \autoref{sec:relbert:the-effect-of-template}.}:
\begin{enumerate}
    \item  Today, I finally discovered the relation between \textbf{[h]} and \textbf{[t]} : \textbf{[h]} is the \texttt{<mask>} of \textbf{[t]}
    \item Today, I finally discovered the relation between \textbf{[h]} and \textbf{[t]} : \textbf{[t]} is \textbf{[h]}'s \texttt{<mask>}
    \item Today, I finally discovered the relation between \textbf{[h]} and \textbf{[t]} : \texttt{<mask>}
    \item I wasn’t aware of this relationship, but I just read in the encyclopedia that \textbf{[h]} is the \texttt{<mask>} of \textbf{[t]}
    \item I wasn’t aware of this relationship, but I just read in the encyclopedia that \textbf{[t]} is \textbf{[h]}’s \texttt{<mask>}
\end{enumerate}
where \texttt{<mask>} is the LM's mask token, and \textbf{[h]} and \textbf{[t]} are slots that are filled with the head word $h$ and tail word $t$ from the given word pair. As a final step, we construct the relation embedding $\mathbf{x}_{(h,t)}$ from  the contextualised representation of the prompt in the LM's output layer. In particular, we have experimented with the following three strategies:
\begin{itemize}
\item We take the contextualised representation of the \texttt{<mask>} token as the relation embedding (\emph{average}).
\item We average the contextualised embeddings across all tokens from the prompt (\emph{mask}).
\item We average the contextualised embeddings across all tokens from the prompt except for the \texttt{<mask>} token (\emph{average w.o. mask}).
\end{itemize}
In the following we explain the training objectives and how the model is trained.

\subsection{Training Objective}\label{sec:relbert:relational-knowledge-distillation-via-language-model-finetuning}
The LM encoder used in RelBERT is initialised from a pre-trained RoBERTa model, which was shown to be more effective than BERT in our previous work \cite{ushio-etal-2021-distilling}. It is then fine-tuned using a contrastive loss, based on the idea that word pairs which belong to the same relation should have a relation embedding that is similar, whereas word pairs belonging to different relations should have embeddings that are further apart. To this end, we assume access to a set of positive training examples $\mathcal{P}_r$ and a set of negative examples $\mathcal{N}_r$, for a number of relations $r\in \mathcal{R}$. In particular, $\mathcal{P}_r$ contains word pairs $(h,t)$ which belong to relation $r$, whereas $\mathcal{N}_r$ contains examples of word pairs which do not. We consider three different loss functions to implement the proposed idea.

\paragraph{Triplet Loss}
The \emph{triplet loss} \cite{schroff2015facenet} relies on training data in the form of triples $(a,p,n)$, where $a$ is called the anchor, $p$ is a positive example, and $n$ is a negative example. The aim of this loss is to ensure that the distance between $a$ and $p$ is smaller, by some margin, than the distance between $a$ and $n$. In our case, the elements $a$, $p$ and $n$ correspond to word pairs, where $a,p\in \mathcal{P}_r$ and $n\in \mathcal{N}_r$ for some relation $r$. Let us write $\bm{x}_a$ for the relation embedding of a word pair $a$. We then have the following loss:
\begin{align}
    L_{{\rm tri}} = \sum_{r\in \mathcal{R}} \sum_{(a,p,n)\in \mathcal{P}_r\times \mathcal{P}_r \times \mathcal{N}_r} \max\big(0, \|\bm{x}_a-\bm{x}_p\| - \|\bm{x}_a - \bm{x}_n\| + \Delta \big)
\end{align}
where $\Delta>0$ is the margin and $\|\cdot\|$ is the $l^2$ norm.

\paragraph{InfoNCE}
Information noise constrastive estimation (InfoNCE) \cite{oord2018representation} addresses two potential limitations of the triplet loss. First, while the triplet loss only considers one negative example at a time, InfoNCE can efficiently contrast each positive example to a whole batch of negative examples. Second, while the triplet loss uses the $l^2$ norm, InfoNCE relies on the cosine similarity, which tends to be better suited for comparing embeddings. The InfoNCE loss can be defined as follows:
\begin{align}
    L_{{\rm nce}} = \sum_{r\in\mathcal{R}} \sum_{(a, p)\in \mathcal{P}_r \times \mathcal{P}_r}  \left(
    - \log \frac{
         \exp{\left( \frac{{\rm cos}(\bm{x}_a, \bm{x}_p)}{\tau} \right)}
        }{\exp{\left( \frac{{\rm cos}(\bm{x}_a, \bm{x}_p)}{\tau} \right)} + \sum_{n \in \mathcal{N}_r} \exp{\left(\frac{{\rm cos}(\bm{x}_a, \bm{x}_n)}{\tau} \right)} 
        }
    \right)
\end{align}
where $\tau$ is a temperature parameter to control the scale of the exponential and cos is the cosine similarity.

\paragraph{InfoLOOB}
Info-leave-one-out-bound (InfoLOOB) \cite{furst2021cloob} is a variant of InfoNCE, in which the positive example is omitted from the denominator. This is aimed at preventing the saturation of the loss value, which can occur with InfoNCE due to dominant positives. Applied to our setting, the loss is as follows:
\begin{align}
    L_{{\rm loob}} =  \sum_{r\in\mathcal{R}} \sum_{(a, p)\in \mathcal{P}_r \times \mathcal{P}_r}  \left(
    - \log \frac{
         \exp{\left( \frac{{\rm cos}(\bm{x}_a, \bm{x}_p)}{\tau} \right)}
        }{\sum_{n \in \mathcal{N}_r} \exp{\left(\frac{{\rm cos}(\bm{x}_a, \bm{x}_n)}{\tau} \right)} 
        }
    \right)
\end{align}

\begin{table}[t]
\centering
\begin{tabular}{lcccc}
\toprule
Dataset                                 & RelSim    & ConceptNet    & NELL              & T-REX \\
\midrule
\#relations (train/val/test)                            & 89/89/-   & 28/18/16      & 31/4/6            & 721/602/24  \\
Average \#positive examples per relation& 14.7/3.7/-& 20,824/66/74  & 177/219/225       & 1,767/529/4\\
Relation Hierarchy                      & True      & False         & False             & False \\
Domain                                  & Concepts  & Concepts      & Named entities    & Named entities \\
\bottomrule
\end{tabular}
\caption{Statistics of the training sets that are considered for RelBERT, including the number of relations, the average number of triples per relation in the training / validation / test sets, and the number of unique positive triples; we also specify whether the relations are organised in a hierarchy, the domain from which the entities are coming.}
\label{tab:relbert:relbert-data-stats}
\end{table}

\subsection{Training Data} \label{sec:relbert:dataset}
As described in \autoref{sec:relbert:relational-knowledge-distillation-via-language-model-finetuning}, to train RelBERT we need positive and negative examples of word pairs belonging to particular relations.
In this section, we describe the four datasets that we considered for training RelBERT. The main properties of these datasets are summarised in \autoref{tab:relbert:relbert-data-stats}.
We now present each dataset in more detail. For each dataset, we have a training and validation spit, which are used for training RelBERT and for selecting the hyperparameters. In addition, for most of the datasets, we also select a test set, which will be used for evaluating the model (see \autoref{analogy-questions}).

\begin{table}[t]
\centering
\begin{tabular}{ll}
\toprule
Relation        &   Examples \\
\midrule
Case Relation   &  [designer, fashions], [preacher, parishioner], [hunter, rifle] \\
Meronym (Part-Whole)      &  [building, wall], [team, player], [movie, scene] \\
Antonym (Contrast)        &  [smooth, rough], [difficult, easy], [birth, death] \\
Space-Time      &  [refrigerator, food], [factory, product], [pool, swimming] \\
Representation  &  [diploma, education], [groan, pain], [king, crown] \\
Hypernym (Class Inclusion) &  [furniture, chair], [furniture, chair], [flower, daisy] \\
Synonym (Similar)         &  [couch, sofa], [sadness, melancholia], [confident, arrogance] \\
Attribute       &  [steel, strong], [glass, shattered], [miser, greed] \\
Non Attribute   &  [empty, full], [incomprehensible, understood], [destitution, abundance] \\
Cause-Purpose   &  [tragedy, tears], [fright, scream], [battery, laptop] \\
\bottomrule
\end{tabular}
\caption{Examples of word pairs from each parent relation category in the RelSim dataset.}
\label{tab:rel-sim-examples}
\end{table}

\paragraph{RelSim}
The Relational Similarity Dataset (RelSim)\footnote{Our preprocessed version of this dataset is available at \url{https://huggingface.co/datasets/relbert/semeval2012_relational_similarity}; the original dataset is available at \url{https://sites.google.com/site/semeval2012task2/download}.} was introduced for SemEval 2012 Task 2 \cite{jurgens-etal-2012-semeval}. It contains crowdsourced judgements about 79 fine-grained semantic relations, which are grouped into 10 parent categories. \autoref{tab:rel-sim-examples} shows word pairs randomly sampled from the highest ranked word pairs in each parent category of RelSim. For each semantic relation, a list of word pairs is provided in RelSim, with each word pair being assigned a prototypicality score\footnote{We used the platinum ratings from the original dataset.}. To convert this dataset into the format that we need for training RelBERT, we consider the 79 fine-grained relations and the 10 parent relations separately. For the fine-grained relations, we choose the 10 most prototypical word pairs, i.e.\ the word pairs with the highest scores, as positive examples, while the 10 lowest ranked word pairs are used as negative examples. For the parent relations, the set of positive examples contains the positive word pairs of each of the fine-grained relations that belong to the parent relation. The negative examples for the parent relations are taken to be the positive examples of the other relations. This is because the parent relations are mutually exclusive, whereas the semantic distinction between the fine-grained relations is often very subtle. From the resulting dataset, we randomly choose 80\% of the word pairs for training, and we keep the remaining 20\% as a validation set.

\paragraph{ConceptNet}\label{sec:relbert:conceptnet}
ConceptNet\footnote{Our preprocesed version of this dataset is available at \url{https://huggingface.co/datasets/relbert/conceptnet_relational_similarity}; the original dataset is available at \url{https://conceptnet.io/}.} \cite{speer-havasi-2012-representing} is a commonsense knowledge graph. It encodes semantic relations between concepts, which can be single nouns or short phrases. The knowledge graph refers to a total of 34 different relations. Since the original ConceptNet contains more than two millions of triples, we employ the version released by \cite{li-etal-2016-commonsense}, where the triples are filtered by their confidence score. We use the \emph{test set} consisting of the 1200 most confident tuples as an evaluation dataset, the \emph{dev1} and \emph{dev2} sets consisting of the next 1200 most confident tuples as our validation set, and the \emph{training set} consisting of 600k tuples as our training set\footnote{The filtered version of ConceptNet is available at \url{https://home.ttic.edu/~kgimpel/commonsense.html}.}. We have disregarded any triples with negated relations such as \emph{NotCapableOf} or \emph{NotDesires}, because they essentially indicate the lack of a relationship. The positive examples for a given relation are simply the word pairs which are asserted to have this relation in the knowledge graph. The negative examples for a given relation are taken to be the positive examples for the other relations, i.e.\ $\mathcal{N}_r = \{(a,b) \in \mathcal{P}_{\hat{r}} | \hat{r} \in \mathcal{R}\setminus \{r\}\}$.

\paragraph{NELL-One}\label{sec:relbert:nell-one}
NELL \cite{mitchell2018never} is a system to collect structured knowledge from web. The authors of \cite{xiong-etal-2018-one} compiled and cleaned up the latest dump file of NELL at the time of publication to create a knowledge graph, called NELL-One\footnote{Our preprocessed version of this dataset is available at \url{https://huggingface.co/datasets/relbert/nell_relational_similarity}; the original dataset is available at \url{https://github.com/xwhan/One-shot-Relational-Learning}} for one-shot relational learning. We employ NELL-One with its original split from \cite{xiong-etal-2018-one}, which avoids any overlap between the relation types appearing in the test set, on the one hand, and the relation types appearing in the training and validation sets, on the other hand. 
Similar as for ConceptNet, the positive examples for a given relation are the word pairs that are asserted to belong to that relation in the training set, whereas the negative examples for a relation are the positive examples of the other relations in the training set.

\paragraph{T-REX}\label{sec:relbert:trex}
T-REX\footnote{Our preprocessed version of this dataset is available at \url{https://huggingface.co/datasets/relbert/t_rex_relational_similarity}; the original dataset is available at \url{https://hadyelsahar.github.io/t-rex/}.} \cite{elsahar-etal-2018-rex} is a knowledge base that was constructed by aligning Wikipedia and Wikidata. It contains a total of 20 million triples, all of which are aligned with sentences from introductory sections of Wikipedia articles. We first remove triples if either their head or tail is not a named entity, which reduces the number of triples from 20,877,472 to 12,561,573, and the number of relations from 1,616 to 1,470. Then, we remove relations with fewer than three triples, as we need at least three triples for each relation type to enable fine-tuning \autoref{sec:relbert:relbert-training}, which reduces the number of triples to 12,561,250, and the number of relations to 1,237. One problem with this dataset is that it contains a number of distinct relations which intuitively have the same meaning. For example, the relations \emph{band} and \emph{music by} both represent ``A song played by a musician''. Therefore, we manually mapped such relations onto the same type. Note that this is useful because the strategy for selecting negative examples when training RelBERT implicitly assumes that relations are disjoint. For the same reason, we manually removed relations that subsume more specific relations. 
For example, the relationship \emph{is a} refers to ``hypernym of'', but T-REX also covers more specific forms of hypernymy such as \emph{fruit of}, \emph{religion}, and \emph{genre}. Another example is \emph{is in}, which models the relation ``located in'', but T-REX also contains finer-grained variants of this relation, such as \emph{town}, \emph{state}, \emph{home field}, and \emph{railway line}. We thus remove triples involving relations such as \emph{is a} and \emph{is in}.
This filtering resulted in a reduction to 12,410,726 triples with 839 relation types. We use this dataset, rather than Wikidata itself, because the fact that a triple is asserted in the introductory section of a Wikipedia article suggests that it expresses salient knowledge. This is important because our aim in fine-tuning RelBERT is to distill relational knowledge from the pre-trained LM itself, rather than to learn the knowledge from the training set. We thus ideally want to limit the training data to word pairs whose relationship is captured by the pre-trained LM. The number of times an entity appears in this dataset can be used as an estimate of the salience of that entity. With this in mind, we removed all triples involving entities that appear less than five times in the dataset, which further reduces the number of triples to 1,616,065. 
To create a test set, we randomly chose 34 relation types and we manually selected around 100 verified triples from those relation types.
The training and validation set is created by splitting the remaining relations 80:20 into a training and validation set. 

\section{Evaluation Tasks}\label{sec:relbert:evaluation-tasks}
We evaluate RelBERT on two relation-centric tasks: analogy questions (unsupervised), and lexical relation classification (supervised). In this section, we describe these tasks and introduce the benchmarks included in our evaluation.

\subsection{Analogy Questions}\label{analogy-questions}
\begin{table}[t]
\centering
\begin{tabular}{lcc}
\toprule
   Dataset &  Avg.\ {\#}Answer Candidates & {\#}Questions \\
\midrule
       SAT &                        5/5 &               - / 374 \\
        U2 &                        4/4 &              24 / 228 \\
        U4 &                        4/4 &              48 / 432 \\
    Google &                        4/4 &              50 / 500 \\
      BATS &                        4/4 &           199 / 1,799 \\
      SCAN &                      72/74 &           178 / 1,616 \\
  NELL-One &                        5/7 &             400 / 600 \\
     T-REX &                      74/48 &             496 / 183 \\
ConceptNet &                      19/17 &         1,112 / 1,192 \\
\bottomrule
\end{tabular}
\caption{Main statistics of the analogy question datasets, showing the  average number of answer candidates, and the total number of questions (validation / test).}
\label{tab:relbert:analogy-questions-stats}
\end{table}

\begin{table}[t]
\centering
\begin{tabular}{lll}\toprule
Dataset                 & Domain                  & Example                                                                 \\ \midrule
SAT                     & College Admission Test  & [beauty, aesthete, pleasure, hedonist]                          \\ \midrule
\multirow{9}{*}{U2}     & Grade4                  & [rock, hard, water, wet]                                        \\
                        & Grade5                  & [hurricane, storm, table, furniture]                            \\
                        & Grade6                  & [microwave, heat, refrigerator, cool]                           \\
                        & Grade7                  & [clumsy, grace, doubtful, faith]                                \\
                        & Grade8                  & [hidden, visible, flimsy, sturdy]                               \\
                        & Grade9                  & [panacea, cure, contagion, infect]                              \\
                        & Grade10                 & [grain, silo, water, reservoir]                                 \\
                        & Grade11                 & [thwart, frustrate, laud, praise]                               \\
                        & Grade12                 & [lie, prevaricate, waver, falter]                               \\ \midrule
\multirow{5}{*}{U4}     & Low Intermediate        & [accident, unintended, villain, evil]                           \\
                        & Low Advanced            & [galleon, sail, quarantine, isolate]                            \\
                        & High Beginning          & [salesman, sell, mechanic, repair]                              \\
                        & High Intermediate       & [classroom, desk, church, pew]                                  \\
                        & High Advanced           & [erudite, uneducated, fervid, dispassionate]                    \\ \midrule
\multirow{4}{*}{BATS}   & Inflectional Morphology & [neat, neater, tasty, tastier]                                  \\
                        & Derivational Morphology & [available, unavailable, interrupted, uninterrupted]            \\
                        & Encyclopedic Semantics  & [stockholm, sweden, belgrade, serbia]                           \\
                        & Lexicographic Semantics & [elephant, herd, flower, bouquet]                               \\ \midrule
\multirow{2}{*}{Google} & Encyclopedic Semantics  & [Canada, dollar, Croatia, kuna]                                 \\
                        & Morphological           & [happy, happily, immediate, immediately]                        \\ \midrule
\multirow{2}{*}{SCAN}   & Metaphor                & [grounds for a building, solid, reasons for a theory, rational] \\
                        & Science                 & [conformance, breeding, adaptation, mating]        \\
\midrule
NELL	&Named Entities&	[Miami Dolphins, Cam Cameron, Georgia Tech, Paul Johnson] \\
\midrule
T-REX	&Named Entities	&[Washington, Federalist Party, Nelson Mandela, ANC] \\
\midrule
ConceptNet&	Concepts	&[bottle, plastic, book, paper]\\
\bottomrule
\end{tabular}
\caption{An example from each domain of the analogy question benchmarks.}
\label{tab:relbert:analogy-questions-example}
\end{table}

Analogy questions are multiple-choice questions, where a given word pair is provided, referred to as the \emph{query pair}, along with a set of candidate answers. Then, the task consists of predicting which is the word pair that is most analogous to the query pair among the given candidate answers. In other words, the task is to find the word pair whose relationship best resembles the relationship between the words in the query \cite{DBLP:conf/ranlp/TurneyLBS03}. 

To solve this task using RelBERT, we simply predict the candidate answer whose relation embedding is most similar to the embedding of the query pair, in terms of cosine similarity.\footnote{More details about how to solve the task with RelBERT can be found in the experimental setting section (\autoref{experimental-setting}).}
For our evaluation, first, we consider the following six analogy question datasets\footnote{Preprocessed versions of the datasets are available at \url{https://huggingface.co/datasets/relbert/analogy_questions} except for SAT, which is not publicly released yet. To obtain the SAT dataset, the author of \cite{DBLP:conf/ranlp/TurneyLBS03} should be contacted.}:

\begin{description}
\item[SAT] The SAT exam is a US college admission test. Turney \cite{DBLP:conf/ranlp/TurneyLBS03} collected a benchmark of 374 word analogy problems, consisting primarily of problems from these SAT tests. Each instance has five candidates. The instances are aimed at college applicants, and are thus designed to be challenging for humans.
\item[U2] Following \cite{boteanu2015solving}, who used word analogy problems from an educational website\footnote{\url{https://www.englishforeveryone.org/Topics/Analogies.html}}, we compiled analogy questions from the same resource\footnote{We use the dataset from the website with permission limited to research purposes.}. They used in particular UNIT 2 of the analogy problems from the website, which have the same form as those from the SAT benchmark, but rather than college applicants, they are aimed at children in grades 4 to 12 from the US school system (i.e.\ from age 9 onwards). We split the dataset into 24 questions for validation and 228 questions for testing. Each question has 4 answer candidates.
\item[U4] We have collected another benchmark from the UNIT 4 problems on the same website that was used for the U2 dataset. These UNIT 4 problems are organised in 5 difficulty levels: high-beginning, low-intermediate, high-intermediate, low-advanced and high-advanced. The low-advanced level is stated to be at the level of the SAT tests, whereas the high-advanced level is stated to be at the level of the GRE test (which is used for admission into graduate schools). The resulting U4 dataset has 48 questions for validation and 432 questions for testing. Each question has 4 answer candidates.
\item[Google] The Google analogy dataset \cite{mikolov-etal-2013-linguistic} has been one of the most commonly used benchmarks for evaluating word embeddings\footnote{The original data is available at \url{https://aclweb.org/aclwiki/Google_analogy_test_set_(State_of_the_art)\#cite_note-1}.}. This dataset contains a mix of semantic and morphological relations such as \textit{capital-of} and \textit{singular-plural}, respectively. The dataset was tailored to the evaluation of word embeddings in a predictive setting. We constructed word analogy problems from the Google dataset by choosing for each correct analogy pair a number of negative examples. To obtain sufficiently challenging negative examples, for each query pair (e.g. \textit{Paris-France}) we extracted three negative instances: 
\begin{enumerate}
    \item two random words from the head of the input relation type (e.g. \textit{Rome-Oslo}); 
    \item two random words from the tail of the input relation type (e.g. \textit{Germany-Canada}); 
    \item a random word pair from a relation type of the same high-level category (i.e.\ semantic or morphological) as the input relation type (e.g. \textit{Argentina-peso}).
\end{enumerate}
The resulting dataset contains 50 validation and 500 test questions, each with 4 answer candidates.
\item[BATS] The coverage of the Google dataset is known to be limiting, and BATS \cite{gladkova-etal-2016-analogy} was developed in an attempt to address its main shortcomings. BATS includes a larger number of concepts and relations, which are split into four categories: lexicographic, encyclopedic, and derivational and inflectional morphology\footnote{The original data is available at \url{https://vecto.space/projects/BATS/}}. We follow the same procedure as for the Google dataset to convert BATS into the analogy question format. The resulting dataset contains 199 validation and 1,799 test questions, each with 4 answer candidates.
\item[SCAN] The relation mapping problem \cite{turney2008latent} is to find a bijective mapping between a set of relations from some source domain and a corresponding set of relations from a given target domain.
SCAN\footnote{The original dataset is available at \url{https://github.com/taczin/SCAN_analogies}.} \cite{czinczoll2022scientific} is an extension of the problems that were collected in \cite{turney2008latent}. Where \cite{turney2008latent} contains 10 scientific and 10 metaphorical domains, SCAN extends them by another 443 metaphorical domains and 2 scientific domains. A single SCAN instance contains a list of the source and the target words ($\bm{a}=[a_1, \dots, a_m]$ and $\bm{b}=[b_1, \dots, b_m]$). We convert such an instance into an analogy question, where the query is $[a_i, a_j]$ and the ground truth is $[b_i, b_j]$ and the negative candidates are $[b_{\hat{i}}, b_{\hat{j}}]$ for $\{(\hat{i}, \hat{j}) \in \{1,\dots, m\} \times \{1,\dots, m\} | (\hat{i}, \hat{j}) \neq (i, j) \}$. This results in 178 and 1,616 questions for the validation and test sets, respectively. The number of answer candidates per question is 74 on average, which makes this benchmark particularly challenging.
\end{description}
In addition, we also converted the validation and test splits of T-REX, NELL and ConceptNet, which were introduced in \autoref{sec:relbert:dataset}, into the format of analogy questions. Note that the validation split is used in our ablation study to compare RelBERT training sets, but not used in the main experiment, where we solve analogy questions in the zero-shot setting. Thus, we do not consider approaches that require validation as well as training data, such as \cite{ushio-etal-2021-bert}.
These analogy questions were constructed by taking two word pairs from the same relation type, one of which is used as the query while the other is used as the correct answer. To create negatives for each positive pair, we take $N$ pairs from each of the other relations. We also add the reversed answer pair to the negative (i.e.\ for the positive pair $(h,t)$ we would add $(t,h)$ as a negative), so the number of the negative pairs is $|\mathcal{R}| \times N + 1$ in each split. 
To create benchmarks with different characteristics, we used $N=2$ for T-REX and $N=1$ for Nell and ConceptNet.

\autoref{tab:relbert:analogy-questions-stats} summarises the main features of the analogy question datasets, and \autoref{tab:relbert:analogy-questions-example} shows an example from each category and dataset.

\subsection{Lexical Relation Classification}\label{sec:lexcal-relation-classification}

\begin{table}[t]
\centering
\begin{tabular}{l lllll}
\toprule
\textbf{}   & \multicolumn{1}{c}{BLESS} & \multicolumn{1}{c}{CogALex} & \multicolumn{1}{c}{EVALution} & \multicolumn{1}{c}{K\&H+N} & \multicolumn{1}{c}{ROOT09} \\ \midrule
Antonym           &                - &      241 / 360 &  1095 / 90 / 415 &                   - &                - \\
Attribute         &    1892 / 143 / 696 &            - &    903 / 72 / 322 &                   - &                - \\
Co-hyponym       &    2529 / 154 / 882 &            - &             - &  18134 / 1313 / 6349 &    2222 / 162 / 816 \\
Event             &    2657 / 212 / 955 &            - &             - &                   - &                - \\
Hypernym          &       924 / 63 / 350 &      255 / 382 &  1327 / 94 / 459 &     3048 / 202 / 1042 &    2232 / 149 / 809 \\
Meronym           &    2051 / 146 / 746 &      163 / 224 &     218 / 13 / 86 &          755 / 48 / 240 &                - \\
Possession &                - &            - &    377 / 25 / 142 &                   - &                - \\
Random            &  8529 / 609 / 3008 &  2228 / 3059 &             - &  18319 / 1313 / 6746 &  4479 / 327 / 1566 \\
Synonym           &                - &      167 / 235 &    759 / 50 / 277 &                   - &                - \\
\bottomrule
\end{tabular}

\caption{Number of instances for each relation type across training / validation / test sets of all lexical relation classification datasets.
\label{tab:relbert:data-stats-lexical-relation-classification}
}
\end{table}

We consider the supervised task of relation classification. This task amounts to classifying word pairs into a predefined set of possible relation types.\footnote{Preprocessed versions of the datasets are available at \url{https://huggingface.co/datasets/relbert/lexical_relation_classification}} To solve this task, we train a multi-layer perceptron (MLP) with one hidden layer, which takes the RelBERT relation embedding of the word pair as input. The RelBERT encoder itself is frozen, since our focus is on evaluating the quality of the RelBERT relation embeddings. We consider the following widely-used multi-class relation classification benchmarks: K\&H+N \cite{necsulescu-etal-2015-reading}, BLESS \cite{baroni-lenci-2011-blessed}, ROOT09 \cite{santus-etal-2016-nine}, EVALution \cite{santus-etal-2015-evalution}, and CogALex-V Subtask 2 \cite{santus-etal-2016-cogalex}. \autoref{tab:relbert:data-stats-lexical-relation-classification} shows the size of the training, validation and test splits for each of these datasets, as well as the kinds of relations they cover. The hyperparameters of the MLP classifier are tuned on the validation split of each dataset. In particular, we tune the learning rate from $[0.001, 0.0001, 0.00001]$ and the hidden layer size from $[100, 150, 200]$. CogALex-V has no validation split, so for this dataset we employ the default configuration of Scikit-Learn \cite{JMLR:v12:pedregosa11a}, which uses a 100-dimensional hidden layer and is optimized using Adam with a learning rate of 0.001.

\section{Experimental Setting}\label{experimental-setting}
In this section, we explain the RelBERT training details (\autoref{sec:relbert:relbert-training}) and we introduce the baselines for analogy questions (\autoref{sec:relbert:baseline-for-analogy-quesionts}) and lexical relation classification (\autoref{sec:relbert:baseline-for-lexical-relation-classification}). Throughout this paper, we rely on the weights that were shared by HuggingFace \cite{wolf-etal-2020-transformers} for all pre-trained LMs. A complete list of the models we used can be found in \ref{app:relbert:hf-name}.

\subsection{RelBERT Training}\label{sec:relbert:relbert-training}
In our experiments, we consider a number of variants of RelBERT, which differ in terms of the pre-trained LM that was used for initialising the model, the loss function \autoref{sec:relbert:relational-knowledge-distillation-via-language-model-finetuning}, and the training data \autoref{sec:relbert:dataset}. In each case, RelBERT is trained for 10 epochs. Moreover, we train one RelBERT model for each of the five prompt templates \autoref{sec:relbert:overall-strategy}. The final model is obtained by selecting the epoch and prompt template that achieved the best performance on the validation split, in terms of accuracy.\footnote{The best template and epoch for each model is specified in \ref{app:hyperparameters}.}  The default configuration for RelBERT is to fine-tune a RoBERTa\textsubscript{BASE} or RoBERTa\textsubscript{LARGE} model using InfoNCE on the RelSim dataset. We will refer to the resulting models as RelBERT\textsubscript{BASE} and RelBERT\textsubscript{LARGE} respectively. The other hyper-parameters are fixed as follows. When using the triplet loss, we set the margin $\Delta$ to 1, the learning rate to 0.00002 and the batch size to 32. When using InfoNCE or InfoLoob, we set the temperature $\tau$ to 0.5, the learning rate to 0.000005 and the batch size to 400. In all cases, we fix the random seed as 0 and we use ADAM \cite{kingma2014adam} as the optimiser. To select the aggregation strategy, as a preliminary experiment, we fine-tuned RelBERT\textsubscript{BASE} with each of the three strategies suggested in \autoref{sec:relbert:relational-knowledge-distillation-via-language-model-finetuning}. As we found that \emph{average w.o.\ mask} achieved the best accuracy on the validation set of RelSim, we used this as the default aggregation strategy.

\subsection{Baselines for Analogy Questions}\label{sec:relbert:baseline-for-analogy-quesionts}
We now introduce the baselines we considered for solving analogy questions.

\paragraph{Latent Relation Analysis} 
Latent Relation Analysis (LRA) \cite{Turney:2005:MSS:1642293.1642475} takes inspiration from the seminal Latent Semantic Analysis (LSA) model for learning document embeddings \cite{DBLP:journals/jasis/DeerwesterDLFH90}. The key idea behind LSA was to apply Singular Value Decomposition (SVD) on a document-term co-occurrence matrix to obtain low-dimensional vector representations of documents. LRA similarly uses SVD to learn relation vectors. In particular, the method also constructs a co-occurrence matrix, where the rows now correspond to word pairs and the columns correspond to lexical patterns. Each matrix entry captures how often the corresponding word pair appears together with the corresponding lexical pattern in a given corpus. To improve the quality of the representations, a PMI-based weighting scheme is used, and the method also counts occurrences of synonyms of the words in a given pair. To solve an analogy question, we can compute the LRA embeddings of the query pair and the candidate pairs, and then select the answer whose embedding is closest to the embedding of the query pair, in terms of cosine similarity. For LRA, we only report the published results from \cite{Turney:2005:MSS:1642293.1642475} for the SAT dataset.

\paragraph{Word Embedding}
Since the introduction of the Word2Vec models \cite{mikolov2013distributed}, word analogies have been a popular benchmark for evaluating different word embedding models. This stems from the observation that in many word embedding models, the relation between two words $A$ and $B$ is to some extent captured by the vector difference of their embeddings. Letting $\mathsf{wv}(A)$ be the word embedding of a word $A$, we can thus learn relation embeddings of the form $\mathbf{x}_{(A,B)} = \mathsf{wv}(B) - \mathsf{wv}(A)$. Using these embeddings, we can solve word analogy questions by again selecting the answer candidate whose embedding is most similar to that of the query pair, in terms of cosine similarity. Since some of the analogy questions include rare words, common word embedding models such as word2vec \cite{mikolov2013distributed} and GloVe \cite{pennington-etal-2014-glove} suffer from out-of-vocabulary issues. We therefore used fastText \cite{bojanowski2016enriching} trained on Common Crawl with subword information, which can handle out-of-vocabulary words by splitting them into smaller chunks of characters\footnote{The embedding model is available at \url{https://fasttext.cc/}.}.

\paragraph{Language Models} 
To solve analogy questions using a pre-trained language model, we proceed as follows. Let $(A,B)$ be the query pair. For each answer candidate $(C,D)$ we construct the sentence ``\textit{A} is to \textit{B} what \textit{C} is to \textit{D}'', following \cite{GPT3}. 
We then compute the perplexity of each of these sentences, and predict the candidate that gives rise to the lower perplexity. The exact computation depends on the type of language model that is considered. For CLMs \cite{radford2018improving,radford2019language,GPT3}, such as those in the GPT family, the perplexity of a sentence ${\boldsymbol s}$ can be computed as follows:
\begin{equation}\label{eq:relbert:PPL}
f({\boldsymbol s}) = \exp \left( - \frac{1}{t}\sum_{j=1}^t \log{ P_{\text{clm}}(s_j | {\boldsymbol s}_{j-1} ) } \right)
\end{equation}
where ${\boldsymbol s}$ is tokenized as $[s_1 ... s_t]$ and
$P_{\text{clm}}(s | {\boldsymbol s})$ is the likelihood from an CLM's next token prediction.
For masked language models (MLMs), such as those in the BERT family \cite{devlin-etal-2019-bert,RoBERTa}, we instead use pseudo-perplexity \cite{salazar-etal-2020-masked}, which is defined as in \eqref{eq:relbert:PPL} but with $P_{\text{mask}}(s_j | {\boldsymbol s}_{\backslash j} )$ instead of $P_{\text{clm}}(s_j | {\boldsymbol s}_{j-1} )$, where ${\boldsymbol s}_{\backslash j} = [s_1 \dots s_{j_1} \text{\textlangle mask\textrangle} s_{j+1} \dots s_t ]$ and $P_{\text{mask}}(s_j | {\boldsymbol s}_{\backslash j})$ is the pseudo-likelihood \cite{wang-cho-2019-bert} that the masked token is $s_j$. Finally, for encoder-decoder LMs (ED LMs) \cite{2020t5,lewis-etal-2020-bart}, we split the template in two parts: the phrase ``\textit{A} is to \textit{B}'' is fed into the encoder, and then we use the decoder to compute the perplexity of the phrase ``\textit{C} is to \textit{D}'', using the probability $P_{\text{clm}}$ of the decoder, conditioned by the encoder.
We compare GPT-2 \cite{radford2019language}, GPT-J \cite{gpt-j}, OPT \cite{zhang2022opt}, OPT-IML \cite{iyer2022opt} as CLMs, BERT \cite{devlin-etal-2019-bert} and RoBERTa \cite{RoBERTa} as MLMs, and T5 \cite{2020t5}, Flan-T5 \cite{https://doi.org/10.48550/arxiv.2210.11416}, Flan-UL2 \cite{tay2023ul2} as ED LMs.

\paragraph{OpenAI Models}
OpenAI\footnote{\url{https://openai.com/}} released a commercial API to provide access to their private in-house models such as GPT-3, GPT-4, and ChatGPT (GPT-3.5-turbo). We have used this API to obtain results for those models. For GPT-3, we use the May 2023 endpoint of davinci, the largest GPT-3 model, and follow the same approach as for the public LMs, as explained above (i.e.\ choose the candidate with the lowest perplexity).
We also include the zero-shot results of GPT-3 that were reported in the original GPT-3 paper \cite{GPT3}, which we refer as GPT-3\textsubscript{original}.
Note that the models that can be accessed via the OpenAI API are subject to be changed every six months, which unfortunately limits the reproducibility of the reported results. For the conversational LMs, i.e.\ ChatGPT and GPT-4, the API does not allow us to compute perplexity scores. We therefore do not include them in our main experiments, but an analysis of these models will be provided in \autoref{sec:relbert:chatgpt-gpt4}.

\subsection{Baselines for Lexical Relation Classification}\label{sec:relbert:baseline-for-lexical-relation-classification}
LexNet \cite{shwartz-dagan-2016-path} and SphereRE \cite{wang-etal-2019-spherere} are the current state-of-the-art (SotA) classifiers on the considered lexical relation classification datasets. Both methods rely on static word embeddings \cite{mikolov2013distributed,bojanowski-etal-2017-enriching}. LexNet trains an LSTM \cite{hochreiter1997long} on the word pair by considering it as a sequence of two words, where each word is mapped to its feature map consisting of a number of lexical features such as part-of-speech and the word embedding. SphereRE employs hyperspherical learning \cite{liu2017deep} on top of the word embeddings, which is to learn a feature map from word embeddings of the word pairs to their relation embeddings, which are distributed over the hyperspherical space. In addition to those SotA methods, we use a simple baseline based on word embeddings. Specifically, we train an MLP with a hidden layer in the same way as explained in \autoref{sec:lexcal-relation-classification}. As possible input representations for this classifier, we consider the concatenation of the word embeddings (\textit{cat}) and the vector difference of the word embeddings (\textit{diff}), possibly augmented with the component-wise product of the word embeddings (\textit{cat+dot} and \textit{diff+dot}), which has been shown to provide improvements in lexical relation classification tasks \cite{vu-shwartz-2018-integrating}. We experiment with word embeddings from GloVE\footnote{The embedding model is available from \url{https://nlp.stanford.edu/projects/glove/}.} \cite{pennington-etal-2014-glove} and fastText\footnote{We use the same embedding model used in \autoref{sec:relbert:baseline-for-analogy-quesionts}.}. Finally, we include the results of pair2vec \cite{joshi-etal-2019-pair2vec}, which is a relation embedding model that was trained by aligning word pair embeddings with LSTM-based encodings of sentences where the corresponding word pairs co-occur.

\begin{table}[!t]
\centering
\begin{tabular}
{@{}l@{\hspace{5pt}}lcccccccccc@{}}
\toprule
&Model &   SAT &    U2 &    U4 &  BATS &  Google &  SCAN &  NELL &  T-REX &  CN &  Average \\
\midrule
\multicolumn{2}{@{}l}{Random}&  20.0 &  23.6 &  24.2 &  25.0 &    25.0 &   2.5 &                        14.3 &                          2.1 &                               5.9  & 15.8\\
\multicolumn{2}{@{}l}{LRA}& \textit{56.4} & -  & -  & -  & -  & -  & -  & -  & -  & - \\
\multicolumn{2}{@{}l}{fastText}& 47.1 & 38.2 & 38.4 &  70.7 &    94.6 &  21.7 &                        59.8 &                         23.0 &                              15.2 & 45.1 \\
\midrule
\multirow{4}{*}{\rotatebox{90}{MLM}}
&BERT\textsubscript{BASE}        &  34.5 &  35.1 &  35.2 &  48.4 &    67.6 &  14.5 &  25.2 &    9.8 &         9.4 &     31.1 \\
&BERT\textsubscript{LARGE}       &  32.9 &  36.0 &  36.6 &  59.4 &    79.8 &  14.1 &  32.0 &   14.2 &        14.0 &     35.4 \\ \cmidrule{2-12}
&RoBERTa\textsubscript{BASE}     &  36.4 &  42.1 &  43.3 &  61.8 &    81.0 &  10.6 &  29.0 &   14.8 &        16.6 &     37.3 \\
&RoBERTa\textsubscript{LARGE}    &  40.6 &  50.4 &  49.8 &  69.9 &    88.8 &  12.1 &  38.2 &   36.1 &        16.7 &     44.7 \\ \midrule
\multirow{18}{*}{\rotatebox{90}{Causal LM}}
&GPT-2\textsubscript{SMALL}      &  32.1 &  38.2 &  36.8 &  43.6 &    56.8 &   5.1 &  28.5 &    7.1 &         8.1 &     28.5 \\
&GPT-2\textsubscript{BASE}       &  34.8 &  42.5 &  40.5 &  58.3 &    75.8 &   7.0 &  43.8 &    6.6 &        11.6 &     35.7 \\
&GPT-2\textsubscript{LARGE}      &  36.1 &  42.1 &  42.6 &  60.1 &    75.4 &   8.8 &  39.0 &   12.6 &        12.8 &     36.6 \\
&GPT-2\textsubscript{XL}       &  36.9 &  43.0 &  44.0 &  62.0 &    80.4 &   8.2 &  40.2 &   11.5 &        12.6 &     37.6 \\ \cmidrule{2-12}
&GPT-J\textsubscript{125M}       &  34.0 &  36.8 &  35.6 &  46.6 &    52.8 &   5.6 &  31.8 &   10.9 &         8.4 &     29.2 \\
&GPT-J\textsubscript{1.3B}       &  36.6 &  42.1 &  42.1 &  63.1 &    77.0 &   8.8 &  47.3 &   13.1 &        11.7 &     38.0 \\
&GPT-J\textsubscript{2.7B}       &  38.5 &  44.7 &  43.5 &  63.8 &    83.0 &   8.8 &  40.0 &   16.9 &        14.1 &     39.3 \\
&GPT-J\textsubscript{6B}         &  45.5 &  48.7 &  47.0 &  67.9 &    87.4 &   9.5 &  43.8 &   20.2 &        14.0 &     42.7 \\
&GPT-J\textsubscript{20B}        &  42.8 &  47.8 &  53.7 &  71.3 &    86.4 &  10.0 &  37.2 &   31.7 &        15.7 &     44.1 \\ \cmidrule{2-12}
&GPT-3\textsubscript{original}*& \textit{53.7} & -  & -  & -  & -  & -  & -  & -  & -  & - \\
&GPT-3\textsubscript{davinci}* & 51.8 & 53.5 & 53.2 & 70.8 & 86.0 & 0.84 & 37.8 & 20.7 & 14.2 & 43.9 \\ \cmidrule{2-12}
&OPT\textsubscript{125M}         &  33.7 &  37.3 &  35.4 &  46.1 &    60.0 &   7.1 &  39.7 &   10.4 &        10.8 &     31.2 \\
&OPT\textsubscript{350M}         &  34.0 &  36.8 &  39.1 &  54.8 &    74.2 &   8.2 &  44.7 &   10.9 &        10.6 &     34.8 \\
&OPT\textsubscript{1.3B}         &  38.5 &  40.4 &  43.5 &  62.8 &    83.4 &   8.8 &  46.3 &   11.5 &        14.8 &     38.9 \\
&OPT\textsubscript{30B}          &  47.1 &  52.2 &  51.9 &  71.5 &    88.2 &  10.5 &  45.2 &   20.8 &        19.1 &     45.2 \\ \cmidrule{2-12}
&OPT-IML\textsubscript{1.3B}     &  41.4 &  42.5 &  44.9 &  63.1 &    82.4 &   8.1 &  42.0 &    7.7 &        14.8 &     38.5 \\
&OPT-IML\textsubscript{30B}      &  48.9 &  50.2 &  49.0 &  70.8 &    88.3 &  10.8 &  44.4 &   23.2 &        17.5 &     44.8 \\
&OPT-IML\textsubscript{M-1.3B} &  40.6 &  40.8 &  44.0 &  64.1 &    85.4 &   9.0 &  45.7 &    8.2 &        15.9 &     39.3 \\
&OPT-IML\textsubscript{M-30B}      &  48.9 &  50.2 &  49.0 &  70.8 &    88.3 &  10.8 &  44.4 &   23.2 &        17.5 &     44.8 \\ \midrule
\multirow{11}{*}{\rotatebox{90}{Encoder Decoder LM}}
&T5\textsubscript{SMALL}         &  28.9 &  32.9 &  30.6 &  41.0 &    55.4 &  17.0 &  38.0 &   15.8 &         5.8 &     29.5 \\
&T5\textsubscript{BASE}          &  26.7 &  34.6 &  39.8 &  43.0 &    41.4 &   9.9 &  27.8 &    6.0 &         7.7 &     26.3 \\
&T5\textsubscript{LARGE}         &  30.2 &  35.5 &  40.0 &  48.9 &    51.4 &  13.7 &  25.2 &   11.5 &         9.1 &     29.5 \\
&T5\textsubscript{3B}            &  34.8 &  35.1 &  35.4 &  41.6 &    45.8 &  10.2 &  25.3 &   20.8 &         9.0 &     28.7 \\
&T5\textsubscript{11B}           &  35.6 &  43.0 &  47.5 &  59.9 &    74.6 &  17.9 &  40.8 &   27.3 &        13.8 &     40.0 \\ \cmidrule{2-12}
&Flan-T5\textsubscript{SMALL}    &  26.7 &  36.4 &  41.7 &  42.7 &    49.0 &  11.6 &  35.5 &   16.9 &         8.3 &     29.9 \\
&Flan-T5\textsubscript{BASE}     &  31.8 &  38.6 &  42.1 &  51.4 &    63.6 &  13.7 &  38.8 &   20.8 &         9.6 &     34.5 \\
&Flan-T5\textsubscript{LARGE}    &  36.1 &  40.8 &  43.5 &  52.6 &    63.8 &  13.4 &  38.0 &   20.8 &        11.3 &     35.6 \\
&Flan-T5\textsubscript{XL}       &  42.0 &  49.6 &  50.7 &  66.8 &    86.8 &  17.6 &  46.0 &   30.6 &        17.0 &     45.2 \\
&Flan-T5\textsubscript{XXL}      &  52.4 &  55.7 &  55.6 &  74.7 &    91.2 &  16.3 &  43.8 &   26.8 &        17.9 &     48.3 \\ \cmidrule{2-12}
&Flan-UL2     &  50.0 &  53.1 &  57.2 &  74.9 &    91.8 &  13.4 &  53.8 &   36.1 &        16.8 &     49.7 \\\midrule
\multicolumn{2}{@{}l}{RelBERT\textsubscript{BASE}}&  59.9 &  59.6 &  57.4 &  70.3 &    89.2 &  25.9 &  62.0 &   \textbf{66.7} &        39.8 &     59.0 \\
\multicolumn{2}{@{}l}{RelBERT\textsubscript{LARGE}}&  \textbf{73.3} &  \textbf{67.5} &  \textbf{63.0} &  \textbf{80.9} &    \textbf{95.2} & \textbf{27.2} &  \textbf{65.8} &   {64.5} &        \textbf{47.5} &     \textbf{65.0}  \\
\bottomrule
\end{tabular}
\caption{The accuracy on each analogy question dataset and the averaged accuracy across datasets, where the best model in each dataset shown in bold. Result in italics were taken from the original paper, and the model with * are private models.}
\label{tab:main-analogy}
\end{table}

\begin{figure}[!ht]
    \centering
    \includegraphics[width=\columnwidth]{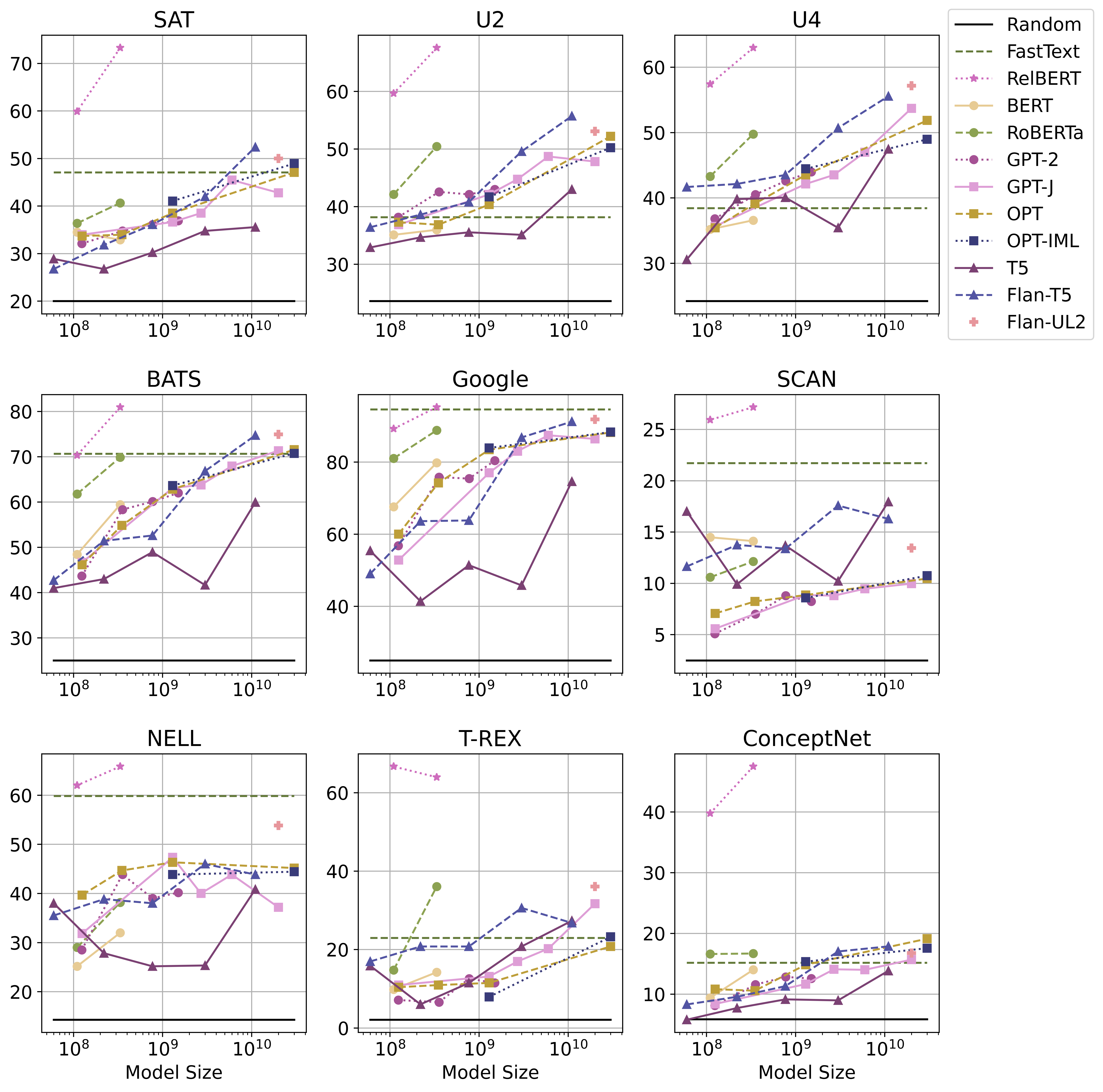}
    \caption{The accuracy on each analogy question dataset in function of the number of parameters in each LM.}
    \label{fig:relbert:main-curve}
\end{figure}

\begin{table}[!t]
\centering
\begin{tabular}{llccccc}
\toprule
Model && BLESS & CogALexV & EVALution & K\&H+N & ROOT09 \\
\midrule
\multicolumn{2}{l}{pair2vec} & 81.7 & 76.9 & 50.5 & 96.9 & 82.9\\ 
\midrule
\multirow{6}{*}{GloVe}  
  &            \textit{cat} & 93.3 & 73.5 & 58.3 & 94.9 & 86.5 \\
  &        \textit{cat+dot} & 93.7 & 79.2 & 57.3 & 95.1 & 89.0 \\
  &   \textit{cat+dot+pair2vec} & 92.6 &    81.1 &    59.6 &  95.7 & 89.4 \\
  &           \textit{diff} & 91.5 &     70.8 &    56.9 &    94.4  &   86.3 \\
  &       \textit{diff+dot} & 92.9 &    78.5 &     57.9 &   94.8 &     88.9 \\
  &  \textit{diff+dot+pair2vec} & 92.2 &     80.2 &     57.4 &     95.5 &     89.4 \\
\midrule
\multirow{6}{*}{fastText}  
  &            \textit{cat} &     92.9 &  72.4 &   57.9 &    93.8 &  85.5 \\
  &        \textit{cat+dot} &     93.2 &    77.4 &    57.8 &    94.0 &    88.5 \\
  &   \textit{cat+dot+pair2vec} &     91.5 &    79.3 &    58.2 &      94.3 &     87.8 \\
  &           \textit{diff} &     91.2 &    70.2 &    55.5 &   93.3 &     86.0 \\
  &       \textit{diff+dot} &       92.9 &    77.8 &     57.4 &      93.6 &     88.9 \\
  &  \textit{diff+dot+pair2vec} &   90.8 &    79.0 &    57.8 &     94.2 &    88.1 \\
  \midrule
\multirow{2}{*}{SotA}  
  & LexNET  & 89.3          & - & 60.0 & 98.5           & 81.3 \\
  & SphereRE& \textbf{93.8} & - & 62.0 & \textbf{99.0}  & 86.1 \\
\midrule
\multicolumn{2}{l}{RelBERT\textsubscript{BASE}}  & 90.0  & 83.7     & {64.2}      & 94.0   & {88.2}   \\
\multicolumn{2}{l}{RelBERT\textsubscript{LARGE}} & 92.0  & \textbf{85.0}     & \textbf{68.4}      & 95.6   & \textbf{90.4}   \\
\bottomrule
\end{tabular}
\caption{Micro F1 score (\%) for lexical relation classification.}
\label{tab:main-classification}
\end{table}

\section{Experimental Results} \label{sec:relbert:experimental-results}
We report the experimental results for the analogy questions benchmarks in \autoref{sec:relbert:result-on-analogy-questions} and for lexical relation classification in \autoref{sec:relbert:result-on-classification}.

\subsection{Results on Analogy Questions}\label{sec:relbert:result-on-analogy-questions}

\autoref{tab:main-analogy} shows the results for each analogy question benchmark in terms of accuracy. We can see that RelBERT substantially outperforms the baselines in all cases, where RelBERT\textsubscript{BASE} is the best for T-REX, and RelBERT\textsubscript{LARGE} is the best for the remaining datasets.
Remarkably, in the case of SAT, none of the pre-trained LMs is able to outperform LRA, a statistical baseline which is almost 20 years old. Moreover, on the Google, SCAN and NELL datasets the LM baselines are outperformed by fastText, a static word embedding model. This clearly shows that LMs struggle with identifying analogies in the zero-shot setting. SCAN and ConceptNet overall emerge as the most challenging benchmarks, which can be largely explained by the large number of answer candidates. Even with the best model, RelBERT\textsubscript{LARGE}, the accuracy is only 27.2\% on SCAN and 47.5\% on ConceptNet. For T-REX, which also involves a large number answer candidates, we can see that the LM baselines are clearly outperformed by RelBERT. Comparing the LM baselines, we find that ED LMs such as Flan-T5\textsubscript{XXL} and Flan-UL2 achieve the best overall results, although CLMs such as GPT-J and OPT\textsubscript{20B} are also competitive among the larger models.

For the LM baselines, unsurprisingly there is a strong correlation between model size and performance. To see this impact more closely, \autoref{fig:relbert:main-curve} plots the accuracy of each LM in function of model size. We can see that the RelBERT models achieve the best result despite being two orders of magnitude smaller than Flan-T5\textsubscript{XXL} and Flan-UL2. Interestingly, RoBERTa usually outperforms the other LM baselines of comparable size, except on NELL and SCAN. This suggests that the strong performance of RelBERT is at least in part due to the use of RoBERTa as the underlying model. Our analysis in \autoref{sec:relbert:the-choice-of-language-model} will provide further support for this hypothesis.
The CLMs (GPT-2, GPT-J, OPT, and OPT-IML) behave rather similarly. They improve as the model size increases, but they are generally worse than the ED LMs and MLMs.
Finally, the graphs in \autoref{fig:relbert:main-curve} make it particularly clear how much the LMs are struggling to compete with fastText in some of the datasets. For example, Flan-T5 and OPT-IML generally outperform fastText only for the largest models, while none of the LMs outperform fastText in SCAN and NELL. Given the superior performance of LMs in many downstream tasks, it is surprising to see LMs underperforming a static word embedding model.
Finally, we can confirm that RelBERT outperforms GPT-3\textsubscript{davinci} in all the datasets. 

\paragraph{Prediction Breakdown}

\begin{table}[!t]
\centering
\begin{tabular}{@{}lc@{\hspace{7pt}}cc@{\hspace{7pt}}c@{\hspace{7pt}}cc@{\hspace{7pt}}c@{}}
\toprule
{}  &  \multicolumn{2}{c}{Google} &  \multicolumn{3}{c}{BATS}&  \multicolumn{2}{c}{SCAN} \\
\cmidrule(l){2-3}\cmidrule(l){4-6}\cmidrule(l){7-8}
{}  &  Encyclopedic &Morphological &  Encyclopedic &  Lexical &  Morphological &  Metaphor &  Science \\
\midrule
Random & 25.0 & 25.0 & 25.0 & 25.0 & 25.0 &2.4 & 2.8\\
fastText &92.6&96.1&71.6&34.0&88.5&18.9&31.9\\
Flan-UL2 &94.4&89.8&68.0&60.2&85.8&11.9&18.8\\
\midrule
RelBERT\textsubscript{BASE}  &                93.0 &   86.3 &        57.8 &     62.9 &           80.3 &      23.4 &     35.0 \\
RelBERT\textsubscript{LARGE} &               98.6 &   92.6 &         71.3 &     72.4 &           90.0 &      24.8 &     35.6 \\
\bottomrule
\end{tabular}
\caption{The accuracy of RelBERT on each domain of three analogy question datasets with random expectation, fastText, and Flan-UL2 as baselines.
}
\label{tab:relbert:domain-accuracy}
\end{table}

\begin{figure}[!t]
    \centering
    \includegraphics[width=0.6\columnwidth]{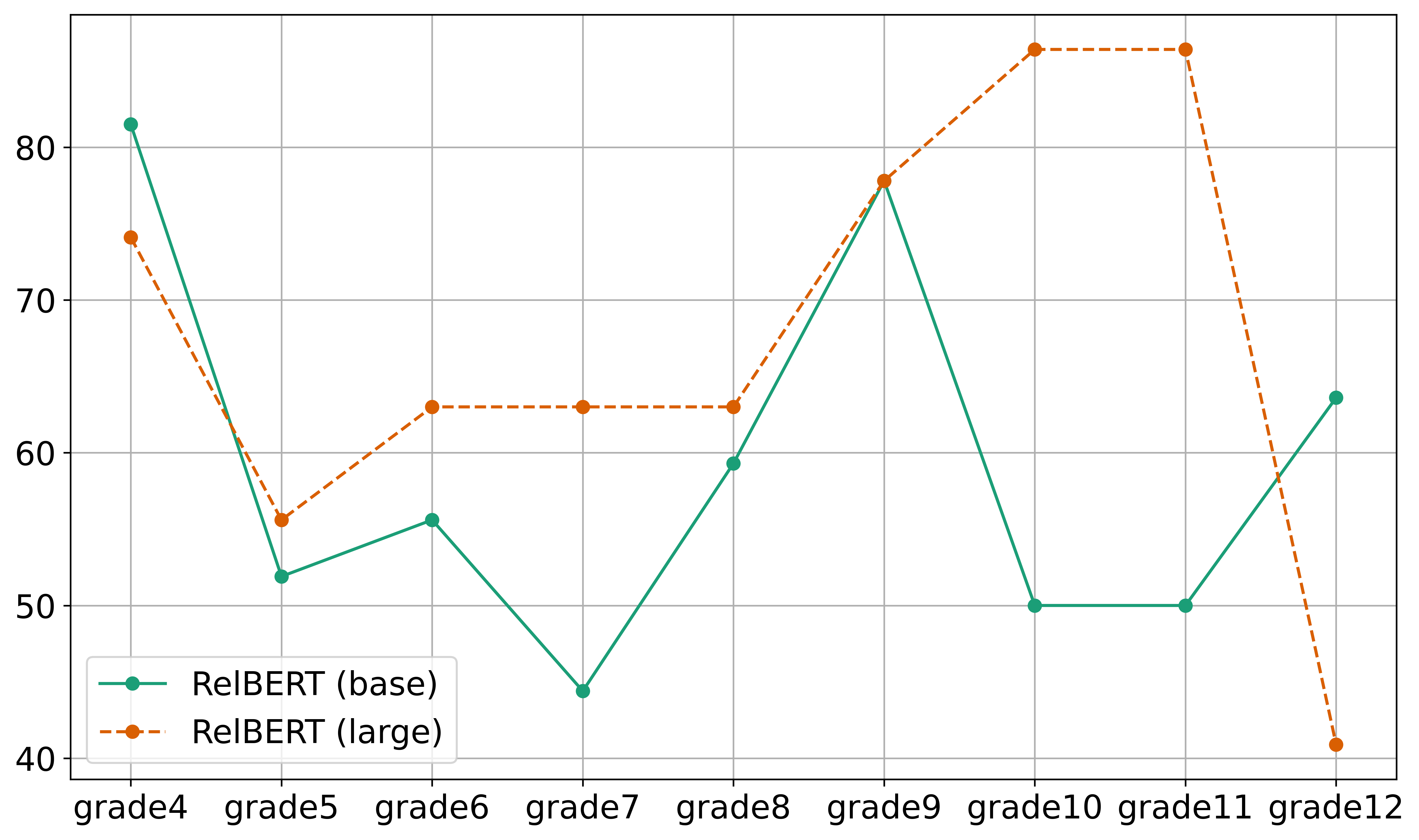}
    \caption{The accuracy of RelBERT for each domain of U2 analogy question.}
    \label{fig:relbert:domain-u2}
\end{figure}

\begin{figure}[!t]
    \centering
    \includegraphics[width=0.6\columnwidth]{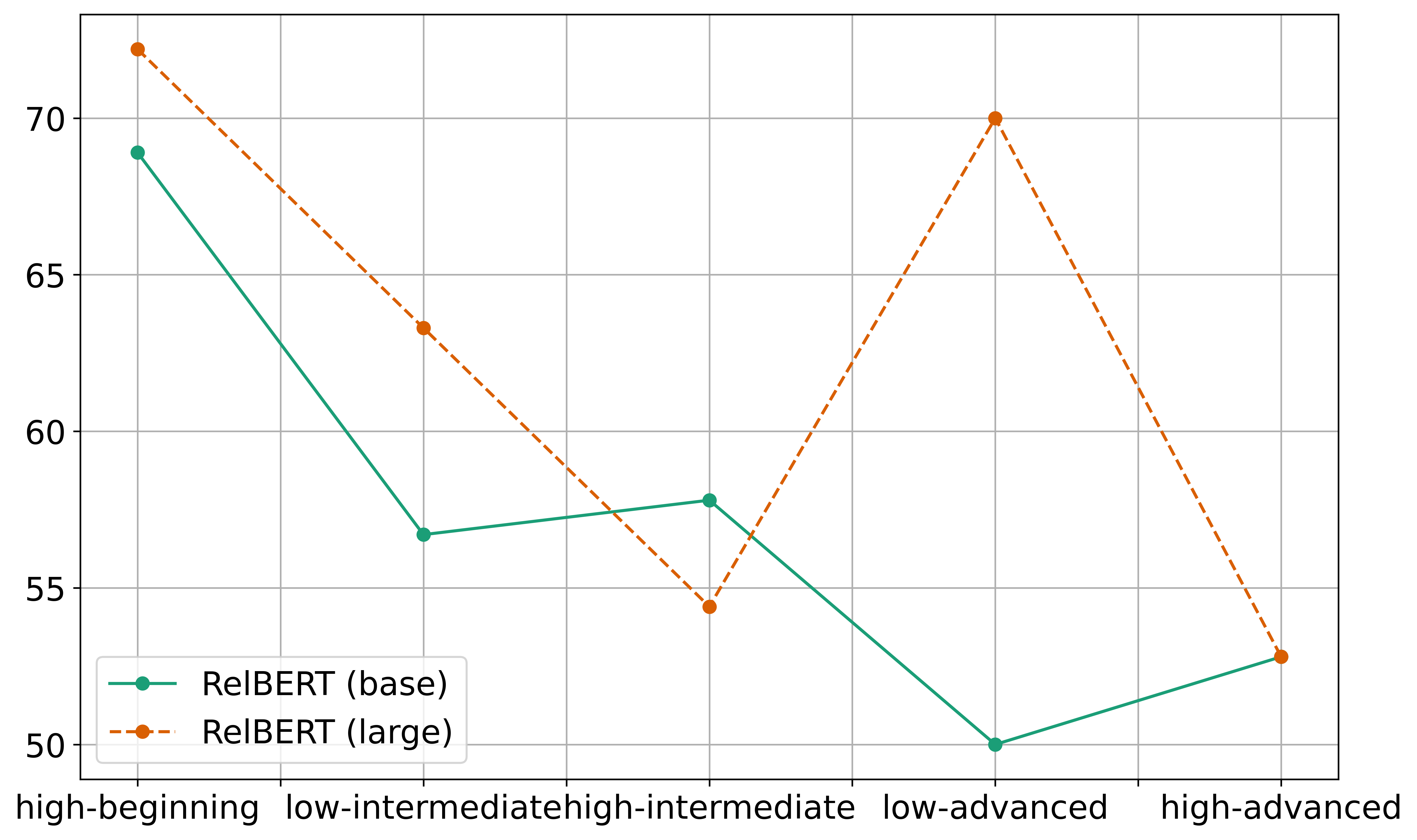}
    \caption{The accuracy of RelBERT for each domain of U4 analogy question.}
    \label{fig:relbert:domain-u4}
\end{figure}

We now analyse the performance of RelBERT on different categories of analogy questions, considering the categories that were listed in \autoref{tab:relbert:analogy-questions-example}. First, the results for some of the categories are shown in \autoref{tab:relbert:domain-accuracy}, along with some baselines. The results show that both RelBERT models can achieve a high accuracy for morphological relationships, despite not being explicitly trained on such relations. This ability appears to increase along with the model size, as RelBERT\textsubscript{LARGE} outperforms RelBERT\textsubscript{BASE} by around 10 percentage points on the morphological relations from BATS, and 6 percentage points on the morphological relations from Google. \autoref{fig:relbert:domain-u2} and \autoref{fig:relbert:domain-u4} show the accuracy along with the difficulty level in U2 and U4. Although we cannot see a clear signal in U2, we can see that models struggle more when the difficulty level is increased in U4, especially for RelBERT\textsubscript{BASE}. Note that the U2 test is designed for children, while U4 is for college students. 
As comparison systems, we included the accuracy breakdown of fastText as a word embedding baseline, and Flan-UL2 as the best LM baseline. RelBERT\textsubscript{LARGE} consistently outperforms  Flan-UL2 in all cases. The performance of fastText is more inconsistent, showing a strong performance on the morphological relations of the Google analogy dataset, as well as the encyclopedic portion of BATS, but performing poorly in the lexical portion of BATS (34.0 compared to RelBERT\textsubscript{Large}'s 72.4) and in the metaphors of SCAN (18.9 compared to 24.8).

\subsection{Lexical Relation Classification}\label{sec:relbert:result-on-classification}
\autoref{tab:main-classification} shows the micro F1 score for the lexical relation classification datasets. We can see that RelBERT\textsubscript{LARGE} is in general competitive with the SotA approaches. For two (EVALution and ROOT09) out of the four lexical relation classification datasets that have SotA results\footnote{These SotA results are reported from the original papers, and thus we could not reproduce in similar conditions.}, RelBERT\textsubscript{LARGE} achieves the best results. Moreover, for these two datasets, even RelBERT\textsubscript{BASE} outperforms the SotA methods. In terms of reproducible word and pair embedding baselines, RelBERT\textsubscript{LARGE} provides better results in all datasets except for BLESS (word embeddings) and K\&H+N (pair2vec). We see a consistent improvement in accuracy when going from RelBERT\textsubscript{BASE} to RelBERT\textsubscript{LARGE}.

\section{Analysis} \label{sec:relbert:analysis}
In this section, we analyse the capability of RelBERT from different aspects. We investigate the generalisation ability of RelBERT for unseen relations in \autoref{sec:relbert:training-data-overlap}. In \autoref{sec:relbert:additional-baselines}, we compare RelBERT with conversational LMs and few-shot learning. Then, we analyse the effect of different design choices in the model architecture in \autoref{sec:relbert:ablation-abalysis}. Finally, in \autoref{sec:relbert:qualitative-analysis} we present a qualitative analysis, where among others we show a visualization of the latent representation space of relation vectors.

\subsection{Generalization Ability of RelBERT}\label{sec:relbert:training-data-overlap}

\begin{table}[!t]
\centering
\begin{tabular}{lcccccc}
\toprule
Dataset\textbackslash Excluded Relation  & Antonym  & Attribute  & Hypernym  & Meronym  & Synonym  &  \emph{Full} \\
\midrule
\emph{BLESS} &&&&&&\\
- Attribute  &     91.6 &       \underline{90.7} &      90.6 &     91.6 &     90.9 &  91.5 \\
- Co-hyponym &     94.6 &       95.3 &      95.5 &     94.0 &     93.8 &  93.5 \\
- Event      &     84.1 &       84.2 &      84.0 &     82.2 &     84.1 &  83.6 \\
- Hypernym   &     92.6 &       93.5 &      \underline{93.5} &     91.3 &     93.1 &  93.1 \\
- Meronym    &     85.7 &       86.8 &      87.5 &     \underline{85.3} &     86.7 &  85.0 \\
- Random     &     92.1 &       92.5 &      92.1 &     91.7 &     91.6 &  91.9 \\
- Average (macro)   &     90.1 &       90.5 &      90.5 &     89.3 &     90.0 &  89.8 \\
- Average (micro)   &     90.6 &       90.9 &      90.8 &     89.9 &     90.3 &  90.2 \\
\midrule
\emph{CogALexV} &&&&&& \\
- Antonym  &     \underline{60.5} &       64.0 &      62.9 &     67.9 &     63.3 &  68.2 \\
- Hypernym &     56.6 &       55.5 &      \underline{56.2} &     56.7 &     56.4 &  59.3 \\
- Meronym  &     70.3 &       69.5 &      65.4 &     \underline{70.3} &     70.5 &  64.5 \\
- Random   &     91.9 &       92.7 &      92.3 &     93.2 &     91.6 &  92.4 \\
- Synonym  &     39.0 &       44.0 &      42.5 &     44.4 &     \underline{42.2} &  45.4 \\
- Average (macro) &     63.7 &       65.1 &      63.9 &     66.5 &     64.8 &  66.0 \\
- Average (micro) &     82.4 &       83.5 &      83.0 &     84.2 &     82.7 &  83.7 \\
\midrule
\emph{EVALution} &&&&&&\\
- Attribute  &     80.7 &       \underline{81.7} &      80.4 &     80.3 &     81.6 &  82.7 \\
- Antonym    &     \underline{72.0} &       74.3 &      73.3 &     75.2 &     73.8 &  73.6 \\
- Hypernym   &     57.7 &       59.3 &      \underline{58.5} &     60.3 &     59.1 &  57.5 \\
- Meronym    &     68.3 &       71.6 &      69.0 &     \underline{64.5} &     66.9 &  68.8 \\
- Possession &     66.7 &       70.3 &      66.4 &     66.0 &     63.5 &  67.4 \\
- Synonym    &     40.6 &       42.9 &      37.4 &     42.9 &     \underline{37.5} &  41.0 \\
- Average (macro)   &     63.6 &       65.7 &      63.7 &     64.3 &     63.0 &  64.5 \\
- Average (micro)   &     64.1 &       65.9 &      64.3 &     65.3 &     63.9 &  65.0 \\
\midrule
\emph{K\&H+N} &&&&&&\\
- Co-hyponym &     95.7 &       96.0 &      94.2 &     96.1 &     94.6 &  95.1 \\
- Meronym    &     63.9 &       63.9 &      57.7 &     \underline{59.8} &     62.4 &  56.7 \\
- Random     &     96.1 &       95.9 &      94.9 &     96.0 &     95.3 &  95.3 \\
- Average (macro)   &     86.7 &       86.8 &      84.0 &     86.1 &     85.6 &  84.5 \\
- Average (micro)   &     95.0 &       95.0 &      93.6 &     95.1 &     94.1 &  94.3 \\
\midrule
\emph{ROOT09} &&&&&&\\
- Co-hyponym &     96.9 &       97.3 &      96.4 &     97.3 &     95.9 &  95.8 \\
- Hypernym   &     80.3 &       80.3 &      \underline{79.0} &     81.8 &     79.2 &  79.5 \\
- Random     &     89.7 &       89.7 &      89.3 &     89.8 &     89.0 &  88.8 \\
- Average (macro)   &     89.0 &       89.1 &      88.2 &     89.7 &     88.0 &  88.0 \\
- Average (micro)   &     89.2 &       89.3 &      88.5 &     89.7 &     88.2 &  88.2 \\
\bottomrule
\end{tabular}
\caption{F1 score for each relation type of all the lexical relation classification datasets from RelBERT\textsubscript{BASE} models fine-tuned on the RelSim without a specific relation. The \emph{Full} model on the right most column is the original RelBERT\textsubscript{BASE} model fine-tuned on full RelSim. The result of the relation type where the model is fine-tuned on RelSim without it, is emphasized by underline.}
\label{tab:training-data-overlap}
\end{table}


RelBERT, when trained on RelSim, achieves competitive results on named entities (i.e.\ Nell-One and T-REX), despite the fact that RelSim does not contain any examples involving named entities. This is one of the most interesting aspects of RelBERT, as it shows that the model learns to infer the relation based on the knowledge from the LM, instead of memorizing the word pairs from the training set. To understand the generalisation ability of RelBERT in more depth, we conduct an additional experiment, where we explicitly exclude a specific relation from RelSim when training RelBERT. Specifically, we train RelBERT on a number of variants of RelSim, where each time a different relation type is excluded. We then test the resulting model on the lexical relation classification datasets. We focus this analysis on the \emph{Antonym}, \emph{Attribute}, \emph{Hypernym}, \emph{Meronym}, and \emph{Synonym} relations, as they are covered by both RelSim (see \autoref{tab:rel-sim-examples}) and at least one of the lexical relation classification datasets (see \autoref{tab:relbert:data-stats-lexical-relation-classification}). We train RelBERT\textsubscript{BASE} with InfoNCE on the different RelSim variants. \autoref{tab:training-data-overlap} shows the results. It can be observed that the performance reduces by at most a few percentage points after removing a given target relation. In some cases, we can even see that the results improve after the removal. Hypernym is covered by all the datasets except K\&H+N, and the largest decrease can be seen for CogALexV, which is around 3 percentage points.  Meronym is covered by all the datasets except ROOT09. After removing the Meronym relation from the training data, the F1 score on meronym prediction increases in three out of four datasets. A similar pattern can be observed for the synonym relation, where the model that was trained without the synonym relation achieves better results than the model trained on the full RelSim dataset. On the other hand, for antonym and attribute, we can see that removing these relations from the training data leads to somewhat lower results on these relations. The average F1 scores over all the relation types are also competitive with, and often even better than those for the full model. These results clearly support the idea that RelBERT can generalise beyond the relation types it is trained on.

\subsection{Additional Baselines}\label{sec:relbert:additional-baselines}
We now analyse the performance of two types of additional models: the conversational LMs from OpenAI in \autoref{sec:relbert:chatgpt-gpt4} and models that rely on few-shot demonstrations in \autoref{sec:relbert:fewshot-learning}.

\subsubsection{ChatGPT and GPT-4} \label{sec:relbert:chatgpt-gpt4}

\begin{table}[!t]
\centering
\begin{tabular}{lcc}
\toprule
         & ChatGPT & GPT-4 \\ \midrule
Prompt 1 & 34.7    & 62.5  \\
Prompt 2 & 45.7    & 79.6  \\ 
\bottomrule
\end{tabular}
\caption{The accuracy on SAT analogy question for ChatGPT and GPT-4 with the different prompts.}
\label{tab:relbert:chatgpt-gpt4}
\end{table}

GPT-4 and ChatGPT are two conversational LMs released by OpenAI\footnote{\url{https://openai.com/}}. As for GPT-3, these models are private and can only be accessed through the OpenAI API. Unlike GPT-3 however, we cannot obtain perplexity scores ({or raw model output that would allow us to} compute perplexity) through the API. Therefore, we instead ask those models directly to choose the best answer candidate, using the following two text prompts: 
\begin{enumerate}
\item 
\begin{quote}
Answer the question by choosing the correct option. Which of the following is an analogy? \\
1) \textit{A} is to \textit{B} what \textit{C}$_1$ is to \textit{D}$_1$ \\
2) \textit{A} is to \textit{B} what \textit{C}$_2$ is to \textit{D}$_2$ \\
3) \textit{A} is to \textit{B} what \textit{C}$_3$ is to \textit{D}$_3$ \\
\dots \\
$\kappa$) \textit{A} is to \textit{B} what \textit{C}$_{\kappa}$ is to \textit{D}$_{\kappa}$ \\
The answer is
\end{quote}
\item 
\begin{quote}
Only one of the following statements is correct. Please answer by choosing the correct option. \\
1) The relation between \textit{A} and \textit{B} is analogous to the relation between \textit{C}$_1$ and \textit{D}$_1$ \\
2) The relation between \textit{A} and \textit{B} is analogous to the relation between \textit{C}$_2$ and \textit{D}$_2$ \\
3) The relation between \textit{A} and \textit{B} is analogous to the relation between \textit{C}$_3$ and \textit{D}$_3$ \\
\dots \\
${\kappa}$) The relation between \textit{A} and \textit{B} is analogous to the relation between \textit{C}$_{\kappa}$ and \textit{D}$_{\kappa}$ \\
The answer is
\end{quote}
\end{enumerate}
where (\textit{A},\textit{B}) is the query word pair, and $[C_i, D_i]_{i=1,\dots,\kappa}$ are the candidate word pairs. We manually parse the outputs returned by the model. As GPT-4 is the most expensive endpoint at the moment, we only report the accuracy on the SAT analogy question dataset. \autoref{tab:relbert:chatgpt-gpt4} shows the result, and we can see that GPT-4 achieves state-of-the-art results with one of the prompts, with ChatGPT being considerably worse. However, the gap between two prompts is more than 15 percentage points, which shows that choosing the right prompt is critical when using GPT-4.

\subsubsection{Few-shot Learning} \label{sec:relbert:fewshot-learning}

\begin{table}[!t]
\centering
\begin{tabular}{lcccccc}
\toprule
Random Seed &     0 &     1 &     2 &     3 &     4 &  Average \\
\midrule
1-shot  &  44.4 &  44.7 &  40.1 &  48.4 &  46.0 &     44.7 \\
5-shots  &  46.0 &  47.1 &  39.8 &  44.9 &  49.5 &     45.5 \\
10-shots &  45.7 &  48.9 &  33.7 &  39.6 &  45.5 &     42.7 \\
\bottomrule
\end{tabular}
\caption{The accuracy of [1, 5, 10]-shots learning with five different random seeds.}
\label{fig:relbert:few-shot}
\end{table}

In the main experiment (\autoref{sec:relbert:experimental-results}), we used the LM baselines in a zero-shot setting. However, recent LLMs often perform better when a few examples are provided as part of the input \cite{GPT3,https://doi.org/10.48550/arxiv.2210.11416}. 
The idea is to provide a few (input,output) pairs at the start of the prompt, followed by the target input. This strategy is commonly referred to as few-shot learning or in-context learning. It is most effective for larger LMs, which can recognize the pattern in the (input,output) pairs and apply this pattern to the target input \cite{https://doi.org/10.48550/arxiv.2210.11416,zhang2022opt,iyer2022opt}. Since RelBERT is fine-tuned on RelSim, for this experiment we provide example pairs to the LM input which are taken from RelSim as well.

We focus on the SAT benchmark and the Flan-T5\textsubscript{XXL} model, which was the best-performing LM on SAT in the main experiments. We consider [1, 5, 10]-shot learning. The demonstrations in each experiment are randomly chosen from the training split of RelSim. We use the same template as for the zero-shot learning, both to describe the examples and to specify the target input. For example, in the 5-shot learning setting, a complete input to the model with five demonstrations of $[\hat{A}_i, \hat{B}_i, \hat{C}_i, \hat{D}_i]_{i=1\dots 5}$ and the target query of $[A, B]$ is shown as below.
\begin{quote}
    $\hat{A}_1$ is to $\hat{B}_1$ what $\hat{C}_1$ is to $\hat{D}_1$ \\
    $\hat{A}_2$ is to $\hat{B}_2$ what $\hat{C}_2$ is to $\hat{D}_2$ \\
    $\hat{A}_3$ is to $\hat{B}_3$ what $\hat{C}_3$ is to $\hat{D}_3$ \\
    $\hat{A}_4$ is to $\hat{B}_4$ what $\hat{C}_4$ is to $\hat{D}_4$ \\
    $\hat{A}_5$ is to $\hat{B}_5$ what $\hat{C}_5$ is to $\hat{D}_5$ \\
    \textit{A} is to \textit{B} what
\end{quote}
We run each experiment for five different random seeds (i.e.\ five different few-shot prompts for each setting). \autoref{fig:relbert:few-shot} shows the results. Somewhat surprisingly, the few-shot models consistently perform worse than the zero-shot model, which achieved an accuracy of 52.4 in the main experiment.

\subsubsection{Multiple-choice Prompt}

\begin{table}[!t]
\centering
\begin{tabular}{lcccc}
\toprule
 &  Flan-T5\textsubscript{XXL} &  Flan-UL2 &  OPT-IML\textsubscript{30B} & OPT-IML\textsubscript{M-30B} \\
\midrule
Analogical Statement & 52.4 & 50.0 & 48.9  & 48.9 \\
Multi-choice QA &                35.8 &             40.6 &                         27.3 &                      31.3 \\
\bottomrule
\end{tabular}
\caption{The accuracy with multiple-choice prompting compared to the vanilla prompting strategy with the analogical statement ($A$ is to $B$ what $C$ is to $D$) on SAT.}
\label{tab:relbert:instruction}
\end{table}

In our main experiment, we compute perplexity separately on each candidate, but the task can be formatted using a multiple-choice question answering prompt as well. Such a prompt provides more information to the LMs, but it requires them to understand the question properly. Following a typical template to solve multiple-choice question answering in the zero-shot setting \cite{GPT3,https://doi.org/10.48550/arxiv.2210.11416}, we use the following text prompt 
\begin{quote}
Which of the following is an analogy? \\
1) \textit{A} is to \textit{B} what \textit{C}$_1$ is to \textit{D}$_1$ \\
2) \textit{A} is to \textit{B} what \textit{C}$_2$ is to \textit{D}$_2$ \\
3) \textit{A} is to \textit{B} what \textit{C}$_3$ is to \textit{D}$_3$ \\
\dots \\
$\kappa$) \textit{A} is to \textit{B} what \textit{C}$_{\kappa}$ is to \textit{D}$_{\kappa}$ \\
The answer is
\end{quote}
where (\textit{A},\textit{B}) is the query word pair, and $[C_i, D_i]_{i=1,\dots,\kappa}$ are the candidate word pairs. 
\autoref{tab:relbert:instruction} shows the accuracy on SAT, for the four best performing LMs on SAT in the main experiment. We can see that the multiple-choice prompt is substantially worse for all the LMs.

\subsection{Ablation Analysis}\label{sec:relbert:ablation-abalysis}
In this section, we analyse how the performance of RelBERT depends on different design choices that were made. We look at the impact of the training dataset in \autoref{sec:relbert:other-training-dataset}; the loss function in \autoref{sec:relbert:the-choice-of-loss-function}; the number of negative samples for the InfoNCE loss in \autoref{sec:the-choice-of-negative-samples}; the base language model in \autoref{sec:relbert:the-choice-of-language-model}; the prompt templates in \autoref{sec:relbert:the-effect-of-template}; and the impact of random variations in \autoref{sec:relbert:the-choice-of-random-seed}. Throughout this section, we use RoBERTa\textsubscript{BASE} for efficiency.

\subsubsection{The Choice of Datasets}
\label{sec:relbert:other-training-dataset}

\begin{table}[!t]
\centering
\begin{tabular}{lcccc}\toprule
Dataset                      & RelSim       & NELL          & T-REX & ConceptNet \\ \midrule
\multicolumn{5}{l}{\textit{Analogy Question}}                                \\
SAT                       & \textbf{59.9}   & 36.1          & 46.8  & 44.9        \\
U2                        & \textbf{59.6}   & 39.9          & 42.5  & 42.5      \\
U4                        & \textbf{57.4}   & 41.0          & 44.0  & 41.0        \\
BATS                      & \textbf{70.3}   & 45.5          & 51.0  & 62.0       \\
Google                    & \textbf{89.2}   & 67.8          & 75.0  & 81.0        \\
SCAN                      & \textbf{25.9}   & 14.9          & 19.6  & 21.8        \\
NELL                      & 62.0            & \textbf{82.5} & 72.2  & 66.2        \\
T-REX                     & 66.7            & 69.9          & \textbf{83.6}  & 44.8        \\
ConceptNet                & \textbf{39.8}   & 10.3          & 18.8  & 22.7       \\\midrule
Average                   & \textbf{59.0}   & 45.3          & 50.4  & 47.4      \\\midrule
\multicolumn{5}{l}{\textit{Lexical Relation Classification}}                  \\
BLESS                     & \textbf{90.0}   & 88.6          & 89.3  & 88.8       \\
CogALexV                  & \textbf{83.7}   & 78.6          & 82.8  & 82.8        \\
EVALution                 & 64.2            & 58.7          & \textbf{64.7}  & 62.8      \\
K\&H+N                     & 94.0           & 94.8          & \textbf{95.2}  & 95.0        \\
ROOT09                    & 88.2            & 86.6          & 88.2  & \textbf{88.8}        \\\midrule
Average                   & 84.0   & 81.5          & \textbf{84.1}  & 83.6       \\
\bottomrule
\end{tabular}
\caption{The results on analogy questions (accuracy) and lexical relation classification (micro F1 score) of RelBERT with different training datasets, where the best result across models in each dataset are shown in bold. }
\label{tab:result-training-data-comparison}
\end{table}

RelSim is relatively small and does not cover named entities, although the RelBERT model trained on RelSim still performed the best on T-REX and NELL-One in the main experiments. Here we present a comparison with a number of alternative training sets, to see whether better results might be possible. We are primarily interested to see whether the performance on NELL and T-REX might be improved by training RelBERT on the training splits of these datasets. We fine-tune RoBERTa\textsubscript{BASE} on three datasets introduced in \autoref{sec:relbert:dataset}:  NELL-One, T-REX and ConceptNet. We use InfoNCE in each case. The results are summarised in \autoref{tab:result-training-data-comparison}. We can see that training RelBERT on RelSim leads to the best results on most datasets, and the best result on average by a large margin. This is despite the fact that RelSim is significantly smaller than the other datasets (see \autoref{tab:relbert:relbert-data-stats}). It is particularly noteworthy that training on RelSim outperforms training on ConceptNet even on the ConceptNet test set, even though ConceptNet contains several relation types that are not covered by RelSim. However, when it comes to the relationships between named entities, and the NELL and T-REX benchmarks in particular, training on RelSim underperforms training on NELL or T-REX. 

\subsubsection{The Choice of Loss Function}\label{sec:relbert:the-choice-of-loss-function}

\begin{table}[!t]
\centering
\begin{tabular}{lccc}
\toprule
Dataset              & Triplet       & InfoNCE           & InfoLOOB      \\\midrule
\multicolumn{4}{l}{\textit{Analogy Question}}                                \\
SAT                  & 54.5          & \textbf{59.9} & 58.8          \\
U2                   & 55.3 & \textbf{59.6}          & 57.5          \\
U4                   & \textbf{58.6} & 57.4          & 56.3          \\
BATS                 & \textbf{72.6}          & 70.3 & 67.6          \\
Google               & 86.4 & \textbf{89.2} & 83.8\\
SCAN                 & \textbf{29.5} & 25.9 & 27.0 \\
NELL                 &\textbf{70.7} & 62.0 & 67.5 \\
T-REX                & 45.4& \textbf{66.7}& 65.6 \\
ConceptNet           & 29.4 & 39.8 & \textbf{40.0}\\ \midrule
Average              & 55.8          & \textbf{59.0} & 58.2         \\ \midrule
\multicolumn{4}{l}{\textit{Lexical Relation Classification}}                  \\
BLESS                & 88.7	& 90.0	& \textbf{91.0}\\
CogALexV             & 80.5	& \textbf{83.7}	& 83.3 \\
EVALution            & \textbf{67.7}	& 64.2	& 65.8 \\
K\&H+N               & 93.1	& 94.0	& \textbf{94.9} \\
ROOT09               & \textbf{90.3}	& 88.2	& 89.3 \\
\midrule
Average              & 84.1	& 84.0	& \textbf{84.9} \\
\bottomrule
\end{tabular}
\caption{The results on analogy questions (accuracy) and lexical relation classification (micro F1 score) with different loss functions, where the best result in each dataset is shown in bold. }
\label{tab:relbert:result-loss-function}
\end{table}

In this section, we compare the performance of three different loss functions for training RelBERT. In particular, we fine-tune RoBERTa\textsubscript{BASE} on RelSim, and we consider the triplet loss and InfoLOOB, in addition to InfoNCE (see \autoref{sec:relbert:relational-knowledge-distillation-via-language-model-finetuning} for more in detail). \autoref{tab:relbert:result-loss-function} shows the result of RelBERT fine-tuned with each of the loss functions. We can see that none of the loss functions consistently outperforms the other. On average, InfoNCE achieves the best results on the analogy questions. The difference with InfoLOOB is small, which is to be expected given that InfoNCE and InfoLOOB are closely related. While the triplet loss performs worse on average, it still manages to achieve the best results in four out of nine analogy datasets. For the relation classification experiments, the results are much closer, with InfoNCE now performing slightly worse than the other loss functions.

\subsubsection{The Choice of the Number of Negative Samples}
\label{sec:the-choice-of-negative-samples}
\begin{table}[!t]
\centering
\begin{tabular}{lccccccccccc}
\toprule
Batch Size &   25  &            50  &            100 &            150 &            200 &            250 &            300 &            350 &            400  & 450 & 500\\
\midrule
\multicolumn{4}{l}{\textit{Analogy Question}}                                \\
SAT               &  56.1 &           59.1 &           53.2 &           55.6 &           56.4 &           57.2 &           57.5 &           58.3 &  \textbf{59.9} &           57.2 &           57.2 \\
U2                &  55.3 &           56.1 &           46.1 &           53.9 &           56.1 &  \textbf{60.1} &           56.1 &           56.6 &           59.6 &           59.6 &           55.7 \\
U4                &  56.5 &           57.4 &           52.5 &           54.9 &           57.2 &  \textbf{59.0} &           57.2 &           54.2 &           57.4 &           58.3 &           56.5 \\
BATS              &  71.7 &           69.6 &           66.9 &           72.3 &           69.7 &           70.9 &           72.0 &  \textbf{73.7} &           70.3 &           69.2 &           70.8 \\
Google            &  88.2 &           87.4 &           78.6 &           88.0 &  \textbf{90.2} &           89.6 &           86.4 &           86.2 &           89.2 &           86.6 &           89.8 \\
SCAN              &  25.0 &           27.2 &           26.2 &           30.9 &           26.5 &           25.3 &           28.8 &  \textbf{31.6} &           25.9 &           25.7 &           25.1 \\
NELL              &  67.0 &           66.5 &           71.5 &  \textbf{77.7} &           66.0 &           63.7 &           75.0 &  \textbf{77.7} &           62.0 &           62.8 &           63.5 \\
T-REX             &  59.6 &           60.7 &           57.9 &           57.9 &           62.3 &           60.1 &           57.9 &           60.1 &           66.7 &  \textbf{69.4} &           62.8 \\
ConceptNet        &  36.9 &           39.3 &           31.1 &           29.4 &           40.5 &           39.6 &           31.0 &           31.4 &           39.8 &           38.9 &  \textbf{42.8} \\
\midrule
Average & 57.4 &           58.1 &           53.8 &           57.8 &           58.3 &           58.4 &           58.0 &           58.9 &  \textbf{59.0} &           58.6 &           58.2 \\
\midrule
\multicolumn{4}{l}{\textit{Lexical Relation Classification}}                  \\
BLESS             &  90.3 &           90.7 &           90.3 &           90.6 &           90.6 &           89.5 &  \textbf{91.3} &           90.6 &           89.6 &           90.4 &           89.7 \\
CogALexV          &  66.0 &  \textbf{67.9} &           65.9 &           62.5 &           65.2 &           64.7 &           65.4 &           64.4 &           65.8 &           65.4 &           63.6 \\
EVALution         &  64.8 &           62.1 &  \textbf{65.0} &           62.9 &           63.6 &           64.6 &           63.8 &           63.2 &           62.9 &           62.8 &           64.6 \\
K\&H+N             &  86.0 &           84.8 &           87.0 &  \textbf{87.5} &           85.3 &           85.2 &           85.2 &           87.2 &           84.6 &           85.4 &           85.1 \\
ROOT09            &  87.7 &           88.8 &           87.7 &           88.6 &  \textbf{89.4} &           88.8 &           88.8 &           88.6 &           87.9 &           88.2 &           88.0 \\
\midrule
Average     &  79.0 &           78.9 &  \textbf{79.2} &           78.4 &           78.8 &           78.6 &           78.9 &           78.8 &           78.2 &           78.4 &           78.2 \\
\bottomrule
\end{tabular}
\caption{The results on analogy questions (accuracy) and lexical relation classification (micro F1 score) with different batch size (negative samples) at InfoNCE, where the best result across models in each dataset is shown in bold. }
\label{tab:result-loss-function-negative}
\end{table}

\begin{figure}[!t]
    \centering
    \includegraphics[width=0.8\columnwidth]{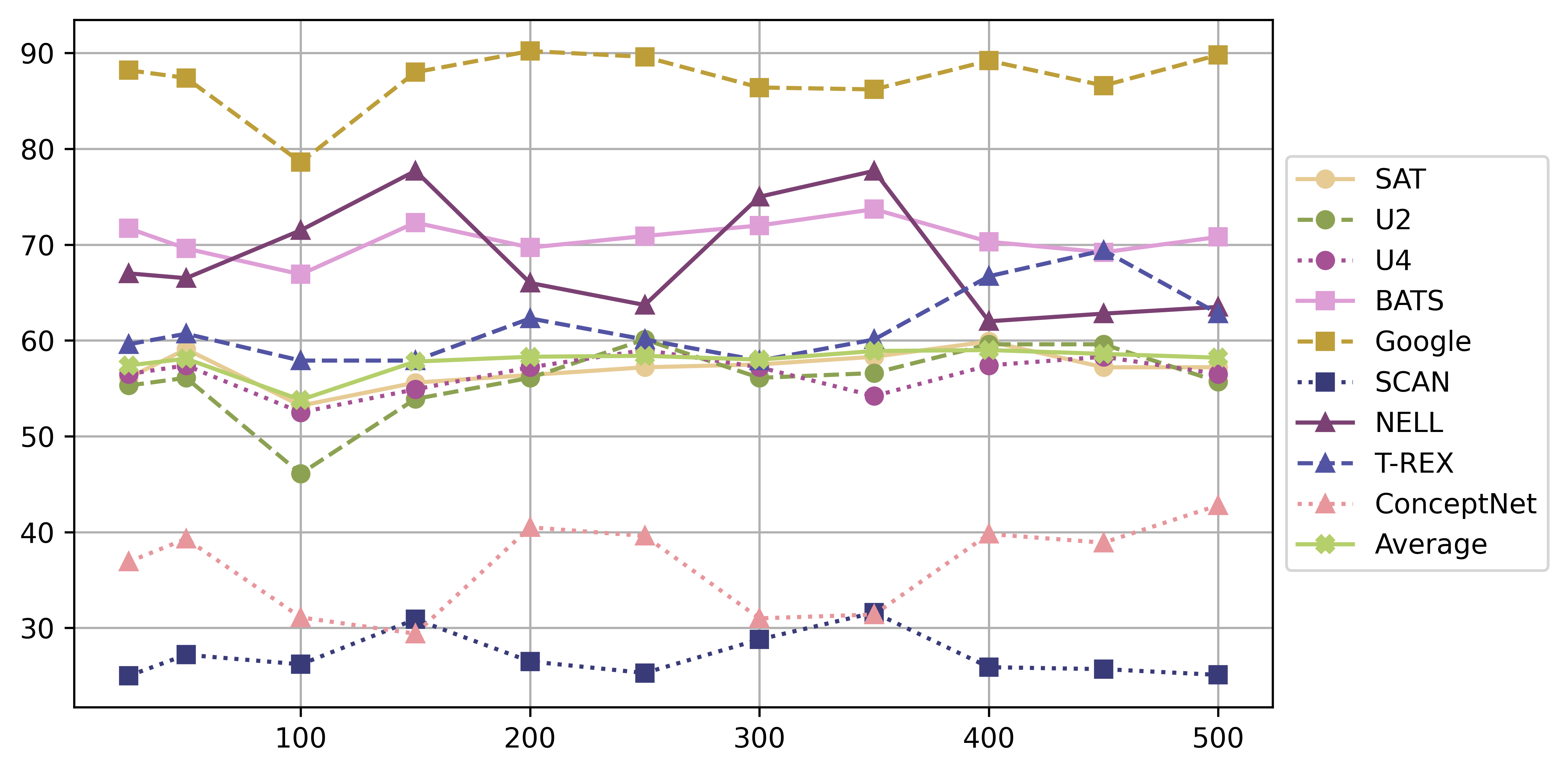}
    \caption{The results on analogy questions in function of the batch size. }
    \label{fig:relbert:negative-sample-ana}
\end{figure}

\begin{figure}[!ht]
    \centering
    \includegraphics[width=0.8\columnwidth]{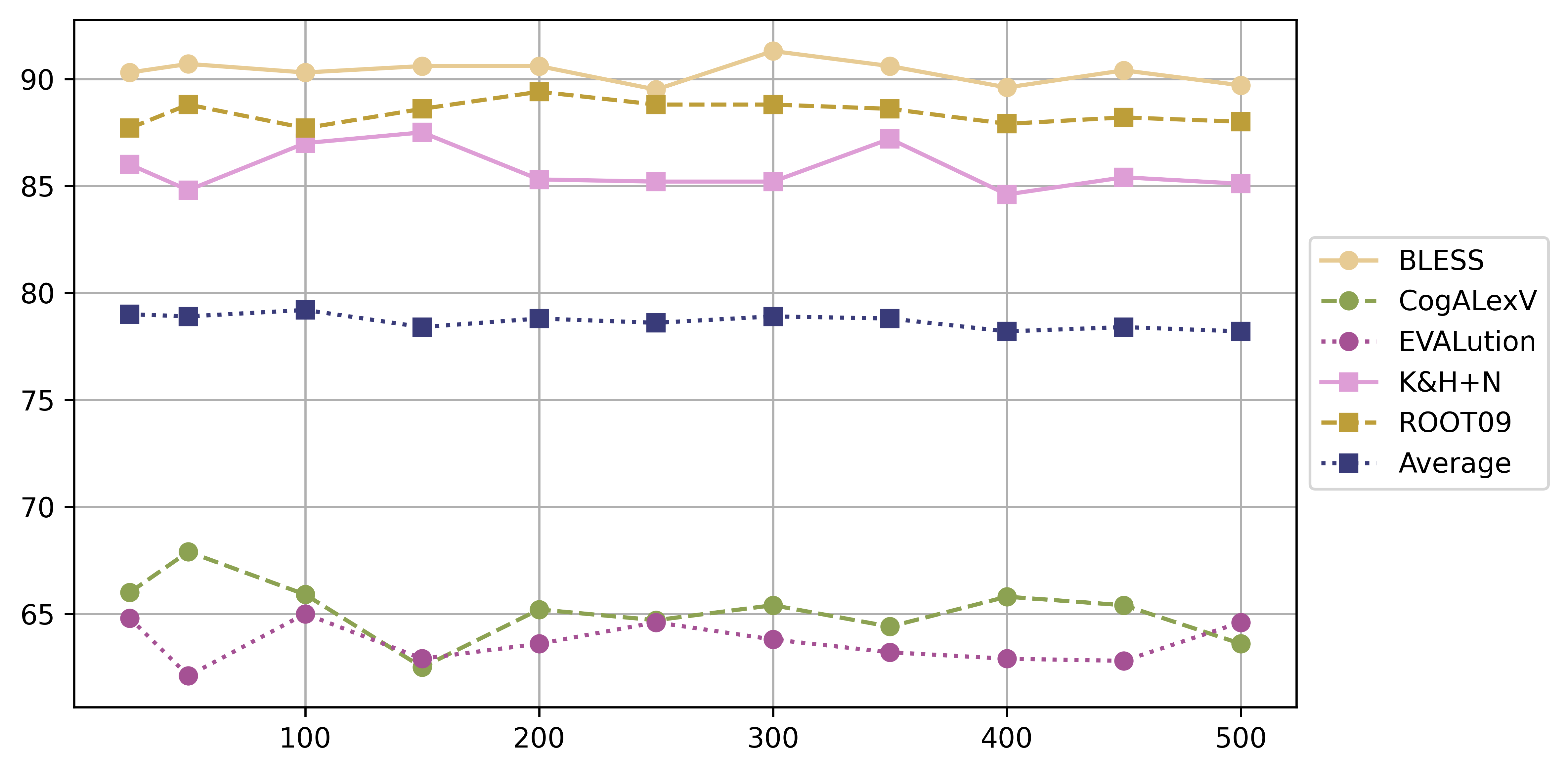}
    \caption{The results on lexical relation classification in function of the batch size. }
    \label{fig:relbert:negative-sample-cls}
\end{figure}

The variant of InfoNCE that we considered for training RelBERT relies on in-batch negative samples, i.e.\ the negative samples for a given anchor pair correspond to the other word pairs that are included in the same batch. The number of negative samples that are considered thus depends on the batch size. In general, using a larger number of negative samples tends to benefit contrastive learning strategies, but it comes at the price of an increase in memory requirement. Here we analyse the impact of this choice, by comparing the results we obtained for different batch sizes. We train RelBERT\textsubscript{BASE} on RelSim with batch sizes from [25, 50, 100, 150, 200, 250, 300, 350, 400, 450, 500], where the batch size 400 corresponds to our main RelBERT\textsubscript{BASE} model. The results are shown in  \autoref{tab:result-loss-function-negative}, and visually illustrated in \autoref{fig:relbert:negative-sample-ana} and \autoref{fig:relbert:negative-sample-cls}.
Somewhat surprisingly, the correlation between batch size and performance is very weak. For analogy questions, there is a weak positive correlation. The Spearman $\rho$ correlation to the batch size for T-REX is 0.6 with p-value 0.047, but in other datasets, correlations are not significant (i.e.\ p-values are higher than 0.05). Indeed, even a batch size of 25 is sufficient to achieve close-to-optimal results. For lexical relation classification, the Spearman correlation is not significant for any of the datasets.

\subsubsection{The Choice of Language Model} \label{sec:relbert:the-choice-of-language-model}

\begin{table}[t]
\centering
\begin{tabular}{lccc}
\toprule
                                         & BERT\textsubscript{BASE} & ALBERT\textsubscript{BASE} & RoBERTa\textsubscript{BASE} \\ \midrule
\multicolumn{4}{l}{\textit{Analogy Question}} \\
SAT                                      & 44.7                      & 40.4                        & \textbf{59.9}                         \\
U2                                       & 36.8                      & 35.5                        & \textbf{59.6}                         \\
U4                                       & 40.0                      & 38.7                        & \textbf{57.4}                         \\
BATS                                     & 54.9                      & 59.2                        & \textbf{70.3}                         \\
Google                                   & 72.2                      & 56.4                        & \textbf{89.2}                         \\
SCAN                                     & 23.7                      & 21.2                        & \textbf{25.9}                         \\
NELL                                     & 56.7                      & 47.7                        & \textbf{62.0}                         \\
T-REX                                    & 49.2                      & 32.8                        & \textbf{66.7}                         \\
ConceptNet                               & 27.1                      & 25.7                        & \textbf{39.8}                         \\ \midrule
Average                                  & 45.0                      & 39.7                        & \textbf{59.0}                         \\
\midrule
\multicolumn{4}{l}{\textit{Lexical Relation Classification}} \\ 
BLESS                                    & \textbf{90.9}                      & 88.0                        & 90.0                         \\
CogALexV                                 & 80.7                      & 78.3                        & \textbf{83.7}                         \\
EVALution                                & 61.8                      & 58.1                        & \textbf{64.2}                         \\
K\&H+N                                    & \textbf{95.5}                      & 92.9                        & 94.0                         \\
ROOT09                                   & \textbf{88.7}                      & 85.6                        & 88.2                         \\ \midrule
Average                                  & 83.5                      & 80.6                        & \textbf{84.0}                        \\
\bottomrule
\end{tabular}
\caption{The results on analogy questions (accuracy) and lexical relation classification (micro F1 score) of RelBERT with different LMs, where the best results across models in each dataset are shown in bold. }
\label{tab:result-language-model-comparison}
\end{table}
Thus far, we have only considered RoBERTa as the base language model for training RelBERT. Here we compare RoBERTa with two alternative choices: BERT \cite{devlin-etal-2019-bert} and ALBERT \cite{lan2019albert}. We compare BERT\textsubscript{BASE}, ALBERT\textsubscript{BASE}, and RoBERTa\textsubscript{BASE}, fine-tuned on RelSim with InfoNCE. \autoref{tab:result-language-model-comparison} shows the result. RoBERTa\textsubscript{BASE} is found to consistently achieve the best results on analogy questions, with a surprisingly large margin. 
RoBERTa\textsubscript{BASE} also achieved the best result, on average, for lexical relation classification, although in this case it only achieves the best results in two out of five datasets. ALBERT consistently has the worst performance, struggling even on the relatively easy Google dataset. These results clearly show that the choice of the LM is of critical importance for the performance of RelBERT.

\subsubsection{The Choice of Prompt Template}\label{sec:relbert:the-effect-of-template}
Our main experiment relies on the five prompt templates introduced in \autoref{sec:relbert:overall-strategy}, where we choose the best among these five templates based on the validation loss. We now analyse the impact of these prompt templates. We focus this analysis on RelBERT\textsubscript{BASE}, i.e.\ RoBERTa\textsubscript{BASE} fine-tuned with InfoNCE on RelSim. For this configuration, the template that was selected based on validation accuracy is
\begin{quote}
Today, I finally discovered the relation between \textbf{[h]} and \textbf{[t]} : \textbf{[h]} is the \texttt{<mask>} of \textbf{[t]}
\end{quote}
We experiment with a number of variations of this template. First, we will see whether the length of the template plays an important role, and in particular whether a similar performance can be achieved with shorter templates. Subsequently we also analyse to what extent the wording of the template matters, i.e.\ whether similar results are possible with templates that are less semantically informative.

\paragraph{The Effect of Length} 

\begin{table}[!t]
\centering
\begin{tabular}{lccccc}
\toprule
Template   & A (Original)             & B             & C             & D             & E    \\ \midrule
\multicolumn{3}{l}{\textit{Analogy Question}} \\      

SAT        & \textbf{59.9} & 58.3          & 56.7 & 59.4          & 46.3          \\
U2         & \textbf{59.6} & 57.9          & 57.9 & 56.6          & 44.7          \\
U4         & 57.4          & 57.6          & 54.9 & \textbf{60.0} & 46.8          \\
BATS       & 70.3          & 69.6          & 70.0 & \textbf{73.9} & 65.6          \\
Google     & 89.2          & 88.0          & 89.4 & \textbf{93.4} & 81.8          \\
SCAN       & 25.9          & 24.8          & 23.9 & \textbf{27.1} & 25.2          \\
NELL       & 62.0          & \textbf{65.5} & 64.2 & \textbf{65.5} & 59.2          \\
T-REX      & \textbf{66.7} & 62.3          & 60.1 & 56.8          & 48.1          \\
ConceptNet & \textbf{39.8} & 39.0          & 37.8 & 39.5          & 32.4          \\
\midrule
Average    & 59.0          & 59.4          & 58.8 & \textbf{61.7} & 51.7          \\
\midrule
\multicolumn{3}{l}{\textit{Lexical Relation Classification}} \\
BLESS      & 89.9          & 89.2          & 89.2 & \textbf{90.5} & 88.2          \\
CogALexV   & 65.7          & 65.3          & 66.7 & \textbf{69.6} & 63.3          \\
EVALution  & \textbf{65.1} & 64.8          & 63.1 & 64.9          & 63.0          \\
K\&H+N      & 85.3          & 85.3          & 83.7 & 86.2          & \textbf{86.9} \\
ROOT09     & 89.1          & 87.6          & 88.9 & \textbf{89.8} & 87.8          \\
\midrule
Average    & 79.0          & 78.4          & 78.3 & \textbf{80.2} & 77.8          \\
\bottomrule
\end{tabular}
\caption{The results on analogy questions (accuracy) and lexical relation classification (micro F1 score) of RelBERT fine-tuned with different length of templates, where the best results across models in each dataset are shown in bold.}
\label{tab:relbert:prompt-length}
\end{table}

\begin{table}[!ht]
\centering
\begin{tabular}{@{}l@{\hspace{4pt}}c@{\hspace{4pt}}c@{\hspace{4pt}}c@{\hspace{4pt}}c@{\hspace{4pt}}c@{\hspace{4pt}}c@{\hspace{4pt}}c@{\hspace{4pt}}c@{\hspace{4pt}}c@{\hspace{4pt}}c@{\hspace{4pt}}c@{}}
\toprule
\multirow{2}{*}{Phrase} & \textit{the}       & \textit{Napoleon}  & \multirow{2}{*}{\textit{football}} & \multirow{2}{*}{\textit{Italy}} & \multirow{2}{*}{\textit{Cardiff}} & \textit{the earth} & \multirow{2}{*}{\textit{pizza}} & \multirow{2}{*}{\textit{subway}} & \multirow{2}{*}{\textit{ocean}} & \textit{Abraham} & \textit{the}      \\
                        & \textit{spaceship} & \textit{Bonaparte} &                                    &                                 &                                   & \textit{science}   &                                 &                                  &                                 & \textit{Lincoln} & \textit{relation} \\
           \midrule
\multicolumn{11}{@{}l}{\textit{Analogy Question}} & \\      
SAT        & 57.2          & 56.7          & 56.1          & 58.0  & 57.2    & 59.6          & 57.0  & 59.6   & 56.7  & \textbf{59.9} & \textbf{59.9} \\
U2         & 55.3          & 58.3          & 56.6          & 55.3  & 57.5    & 58.3          & 55.7  & 57.0   & 57.5  & 56.1          & \textbf{59.6} \\
U4         & 56.9          & \textbf{58.1} & 56.5          & 56.2  & 55.1    & 56.9          & 56.7  & 55.3   & 55.8  & 57.2          & 57.4          \\
BATS       & \textbf{71.2} & 69.1          & 69.5          & 68.8  & 69.5    & 69.9          & 68.5  & 69.8   & 68.9  & 69.5          & 70.3          \\
Google     & 87.2          & 85.4          & 87.6          & 85.2  & 86.8    & \textbf{89.2} & 85.6  & 87.6   & 86.0  & \textbf{89.2} & \textbf{89.2} \\
SCAN       & 25.6          & \textbf{26.8} & 25.7          & 26.1  & 26.2    & 22.6          & 25.8  & 26.6   & 25.6  & 24.6          & 25.9          \\
NELL       & 64.8          & 61.7          & 63.8          & 63.5  & 60.0    & 64.7          & 65.8  & 63.3   & 63.3  & \textbf{66.2} & 62.0          \\
T-REX      & 63.4          & 51.9          & 57.4          & 59.6  & 55.7    & 56.3          & 59.0  & 60.1   & 60.7  & 60.7          & \textbf{66.7} \\
ConceptNet & 39.6          & 39.4          & 38.5          & 36.9  & 37.9    & 39.3          & 38.5  & 38.8   & 38.8  & 38.3          & \textbf{39.8} \\ \midrule
Average    & 57.9          & 56.4          & 56.9          & 56.6  & 56.2    & 57.4          & 57.0  & 57.6   & 57.0  & 58.0          & \textbf{59.0} \\ \midrule
\multicolumn{11}{@{}l}{\textit{Lexical Relation Classification}} & \\   
BLESS      & 89.9          & 89.7          & 90.0          & 90.0  & 89.2    & \textbf{91.4} & 89.7  & 89.7   & 90.6  & 89.3          & 89.9          \\
CogALexV   & 63.4          & 64.7          & \textbf{66.5} & 65.7  & 66.3    & 65.3          & 66.0  & 65.3   & 65.3  & 64.1          & 65.7          \\
EVALution  & 63.6          & 63.7          & 64.0          & 63.8  & 62.4    & 63.5          & 64.5  & 63.3   & 63.5  & 64.0          & \textbf{65.1} \\
K\&H+N      & 84.8          & 86.2          & \textbf{86.4} & 85.9  & 85.3    & 84.8          & 84.7  & 84.6   & 85.1  & 85.0          & 85.3          \\
ROOT09     & 88.5          & 88.9          & 89.4          & 89.0  & 89.2    & 88.7          & 89.1  & 89.5   & 88.9  & \textbf{89.6} & 89.1          \\ \midrule
Average    & 78.0          & 78.6          & 79.3          & 78.9  & 78.5    & 78.7          & 78.8  & 78.5   & 78.7  & 78.4          & \textbf{79.0} \\
\bottomrule
\end{tabular}
\caption{The results on analogy questions (accuracy) and lexical relation classification (micro F1 score) of RelBERT fine-tuned with random phrase to construct the template, where the best results across models in each dataset are shown in bold.}
\label{tab:relbert:prompt-random}
\end{table}

We start from the best template chosen for RelBERT\textsubscript{BASE}, and shorten it while preserving its meaning as much as possible. Specifically, we considered the following variants:
\begin{enumerate}[label=\Alph*.]
    \item Today, I finally discovered the relation between \textbf{[h]} and \textbf{[t]} : \textbf{[h]} is the \texttt{<mask>} of \textbf{[t]}
    \item I discovered the relation between \textbf{[h]} and \textbf{[t]}: \textbf{[h]} is the \texttt{<mask>} of \textbf{[t]}	
    \item the relation between \textbf{[h]} and \textbf{[t]}: \textbf{[h]} is the \texttt{<mask>} of \textbf{[t]}
    \item I discovered: \textbf{[h]} is the \texttt{<mask>} of \textbf{[t]}
    \item \textbf{[h]} is the \texttt{<mask>} of \textbf{[t]}
\end{enumerate}
For each of the templates, we fine-tune RoBERTa\textsubscript{BASE} with InfoNCE on RelSim.  The results are summarised in \autoref{tab:relbert:prompt-length}. We find that template D outperforms the original template A on average, both for analogy questions and for lexical relation classification. In general, we thus find no clear link between the length of the template and the resulting performance, although the shortest template (template E) achieves by far the worst results. This suggests that, while longer templates are not necessarily better, using templates which are too short may be problematic.

\paragraph{The Effect of Semantics} 
We now consider variants of the original template in which the anchor phrase ``the relation'' is replaced by a semantically meaningful distractor, i.e.\ we consider templates of the following form:
\begin{quote}
Today, I finally discovered \texttt{<semantic phrase>} between \textbf{[h]} and \textbf{[t]} : \textbf{[h]} is the \texttt{<mask>} of \textbf{[t]}    
\end{quote}
where \texttt{<semantic phrase>} is a placeholder for the chosen anchor phrase. We randomly chose 10 phrases (four named entities and six nouns) to play the role of this anchor phrase. For each of the resulting templates, we fine-tune RoBERTa\textsubscript{BASE} with InfoNCE on the RelSim dataset.
\autoref{tab:relbert:prompt-random} shows the result. We can see that the best results are obtained with the original template, both for analogy questions and for lexical relation classification. Nevertheless, the difference in performance is surprisingly limited.
The largest decrease is 2.8 in the average for analogy questions and 1.0 in the average for lexical relation classification, which is smaller than the differences we observed when using the shortest template, or when changing the LM \autoref{sec:relbert:the-choice-of-language-model} or the loss function \autoref{sec:relbert:the-choice-of-loss-function}.

\subsubsection{The Choice of Random Seed}\label{sec:relbert:the-choice-of-random-seed}

\begin{table}[!t]
\centering
\begin{tabular}{lccclcccl}
\toprule
      & \multicolumn{4}{c}{RelBERT\textsubscript{BASE}} & \multicolumn{4}{c}{RelBERT\textsubscript{LARGE}} \\
      \cmidrule(l){2-5}\cmidrule(l){6-9}
Random Seed& 0       & 1       & 2       & Average           & 0        & 1        & 2       & Average          \\\midrule
\multicolumn{9}{l}{\textit{Analogy Question}} \\ 
SAT        & 59.9    & 54.3    & 55.9    & 56.7 $\pm$2.9     & 68.2     & 68.4     & 71.9    & 69.5 $\pm$2.1    \\
U2         & 59.6    & 51.8    & 55.7    & 55.7 $\pm$3.9     & 67.5     & 65.4     & 68.0    & 67.0 $\pm$1.4    \\
U4         & 57.4    & 53.7    & 54.4    & 55.2 $\pm$2.0     & 63.9     & 65.0     & 66.4    & 65.1 $\pm$1.3    \\
BATS       & 70.3    & 65.3    & 67.3    & 67.6 $\pm$2.5     & 78.3     & 79.7     & 80.4    & 79.5 $\pm$1.1    \\
Google     & 89.2    & 79.4    & 85.6    & 84.7 $\pm$5.0     & 93.4     & 93.4     & 94.8    & 93.9 $\pm$0.8    \\
SCAN       & 25.9    & 25.9    & 23.3    & 25.1 $\pm$1.5     & 25.9     & 27.0     & 29.1    & 27.4 $\pm$1.6    \\
NELL       & 62.0    & 71.0    & 63.5    & 65.5 $\pm$4.8     & 60.5     & 67.3     & 66.3    & 64.7 $\pm$3.7    \\
T-REX      & 66.7    & 55.2    & 46.4    & 56.1 $\pm$10.1    & 67.8     & 65.0     & 63.4    & 65.4 $\pm$2.2    \\
ConceptNet & 39.8    & 27.4    & 30.5    & 32.6 $\pm$6.4     & 43.3     & 44.8     & 48.7    & 45.6 $\pm$2.8    \\\midrule
Average    & 59.0    & 53.8    & 53.6    & 55.5 $\pm$3.1     & 63.2     & 64.0     & 65.4    & 64.2 $\pm$1.1    \\\midrule
\multicolumn{9}{l}{\textit{Lexical Relation Classification}} \\ 
BLESS      & 90.0    & 91.4    & 90.7    & 90.7 $\pm$0.7     & 91.5     & 92.4     & 91.7    & 91.9 $\pm$0.5    \\
CogALexV   & 83.7    & 81.5    & 81.1    & 82.1 $\pm$1.4     & 84.9     & 86.5     & 86.4    & 85.9 $\pm$0.9    \\
EVALution  & 64.2    & 63.3    & 63.1    & 63.5 $\pm$0.6     & 66.9     & 69.0     & 67.8    & 67.9 $\pm$1.0    \\
K\&H+N     & 94.0    & 94.7    & 94.3    & 94.3 $\pm$0.4     & 95.1     & 95.3     & 95.7    & 95.3 $\pm$0.3    \\
ROOT09     & 88.2    & 88.3    & 89.5    & 88.7 $\pm$0.7     & 89.2     & 89.5     & 91.5    & 90.1 $\pm$1.3    \\\midrule
Average    & 84.0    & 83.8    & 83.7    & 83.9 $\pm$0.1     & 85.5     & 86.5     & 86.6    & 86.2 $\pm$0.6    \\\bottomrule
\end{tabular}
\caption{Result of RelBERT\textsubscript{BASE} and RelBERT\textsubscript{LARGE} with three runs with different random seed, and the average and the standard deviation in each dataset.}
\label{fig:relbert:random}
\end{table}

In this section, we investigate the stability of RelBERT training, by comparing the results we obtained for different random seeds. We use a fixed random seed of 0 as default in the main experiments. Here we include results for two other choices of the random seed. We train both of RelBERT\textsubscript{BASE} and RelBERT\textsubscript{LARGE}. However, different from the main experiments, for this analysis we reduce the batch size from 400 to 100 for RelBERT\textsubscript{LARGE}, to reduce the computation time. \autoref{fig:relbert:random} shows the result. We observe that the standard deviation is higher for RelBERT\textsubscript{BASE} than  for RelBERT\textsubscript{LARGE}. For example, the accuracy on T-REX  differs from 46.4 to 66.7 for RelBERT\textsubscript{BASE}, while only ranging between 63.4 and 67.8 for RelBERT\textsubscript{LARGE}. We can also see that there is considerably less variation in performance for lexical relation classification, compared to analogy questions.

\subsection{Qualitative Analysis} \label{sec:relbert:qualitative-analysis}

\begin{table}[!t]
\centering
\begin{tabular}{lllll}
\toprule
& & AtLocation                      & CapableOf                        & IsA                         \\ \midrule
\multirow{6}{*}{\rotatebox{90}{RelBERT\textsubscript{LARGE}}} & \multirow{3}{*}{1}& child:school, animal:zoo,        & dog:bark, cat:hunt mouse         & baseball:sport, sushi:food, \\
                                                              && prisoner:jail, fish:aquarium,    & lawyer:settle lawsuit,           & yo-yo:toy, dog:mammal,      \\
                                                              & & librarian:library                & pilot:land plane                 & spanish:language            \\ \cmidrule(l){2-5} 
                                                              &\multirow{3}{*}{2}& computer:office, book:shelf,     & knife:spread butter,             & canada:country,             \\
                                                              && coat:closet, food:refrigerator,  & clock:tell time, comb:part hair, & california:state,           \\
                                                              && paper clip:desk                  & match:light fire                 & san francisco:city          \\ \midrule
\multirow{6}{*}{\rotatebox{90}{RelBERT\textsubscript{BASE}}}  &\multirow{3}{*}{1}& animal:zoo, elephant:zoo,        & dog:bark, student:study,         & baseball:sport,             \\
                                                              && student:classroom,               & ball:bounce, tree:grow,          & soccer:sport,               \\
                                                              && student:school                   & bomb:explode                     & chess:game                  \\ \cmidrule(l){2-5} 
                                                              &\multirow{3}{*}{2}& computer:office, coat:closet,    & plane:fly, clock:tell time,      & dog:mammal, rose:flower,    \\
                                                              && mirror:bedroom, notebook:desk,   & computer:compute,                & violin:string instrument,   \\
                                                              && food:refrigerator                & knife:spread butter              & fly:insect, rice:food       \\ \midrule
\multirow{6}{*}{\rotatebox{90}{fastText}}                     &\multirow{3}{*}{1}& fish:water, child:school,        & cat:hunt mouse, cat:drink water, &                            \\
                                                              && bookshelf:library, feather:bird, & dog:guide blind person,          &                             \\
                                                              && computer:office                  & dog:guard house                  &                             \\ \cmidrule(l){2-5} 
                                                              &\multirow{3}{*}{2}& animal:zoo, elephant:zoo,        & knife:spread butter,             &                            \\
                                                              && lion:zoo, weasel:zoo,            & turtle:live long time,           &                             \\
                                                              && tiger:zoo                        & magician:fool audience           &                             \\ \bottomrule
\end{tabular}
\caption{Examples of word pairs in the clusters obtained by HDBSCAN for different relation embeddings. For the \emph{IsA} relation, all word pairs were clustered together in the case of fastText.}
\label{tab:relbert:cluster}
\end{table}

\begin{figure}[!t]
    \centering
    \begin{subfigure}[b]{\columnwidth}
        \centering
        \includegraphics[width=\columnwidth]{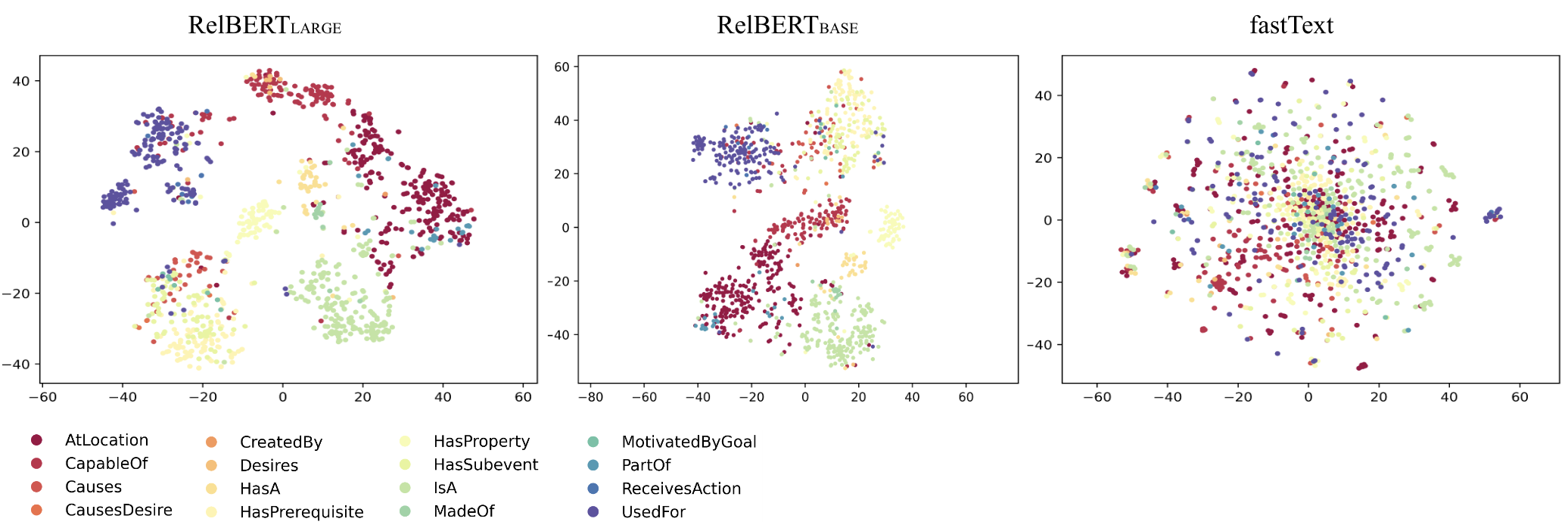}
        \caption{ConceptNet}
        \label{fig:relbert:latent-cn}
    \end{subfigure}     
    \begin{subfigure}[b]{\columnwidth}
        \centering
        \includegraphics[width=\columnwidth]{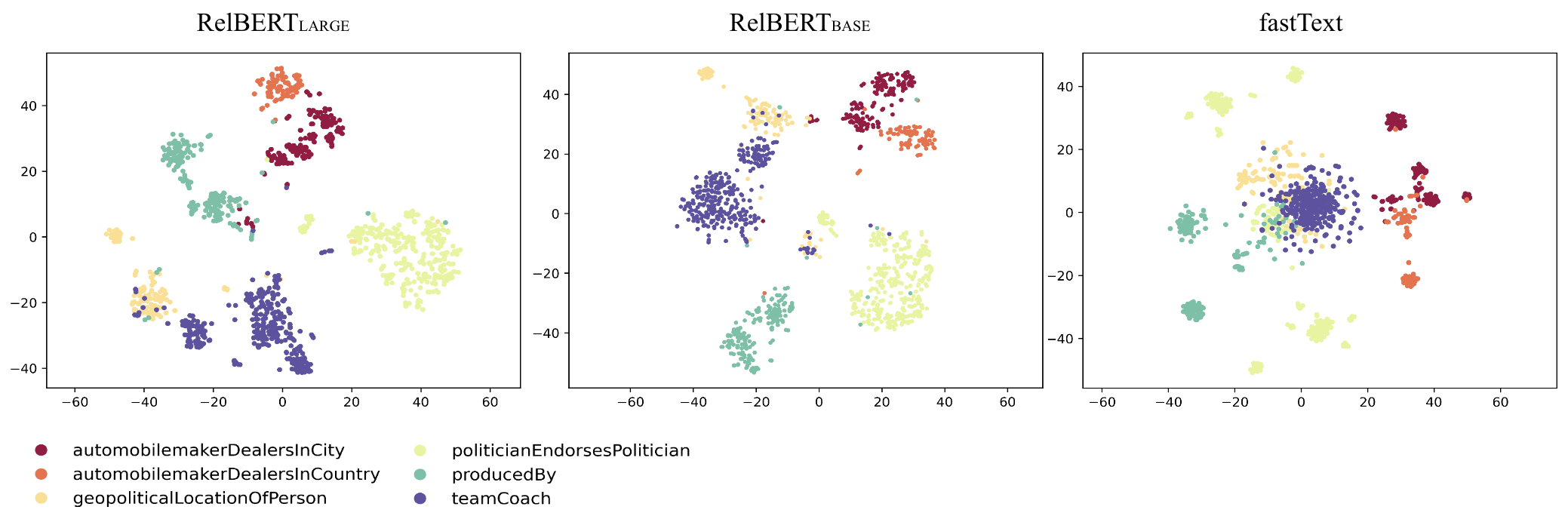}
        \caption{NELL-One}
        \label{fig:relbert:latent-nell}
    \end{subfigure}
    \caption{The tSNE 2-dimension visualization of relation embeddings over the test set of (a) ConceptNet and (b) NELL-One. Colours reflect the relation type of the embedded word pairs.}
    \label{fig:relbert:latent}
\end{figure}

We qualitatively analyse the latent space of the RelBERT relation vectors. For this analysis, we focus on the test splits of the ConceptNet and NELL-One datasets. We compute the relation embeddings of all the word pairs in these datasets using RelBERT\textsubscript{BASE} and RelBERT\textsubscript{LARGE}. As a comparison, we also compute relation embeddings using fastText, in the same way as in \autoref{sec:relbert:baseline-for-analogy-quesionts}. First, we visualize the relation embeddings, using tSNE \cite{TSNE} to map the embeddings to a two-dimensional space. \autoref{fig:relbert:latent} shows the resulting two-dimensional relation embeddings. The plots clearly show how the different relation types are separated much more clearly for RelBERT than for fastText. For ConceptNet, in particular, we can see that the fastText representations are mixed together. Comparing RelBERT\textsubscript{LARGE} and RelBERT\textsubscript{BASE}, there is no clear difference for NELL-One. For ConceptNet, we can see that RelBERT\textsubscript{LARGE} leads to clusters which are somewhat better separated.

Second, we want to analyse whether RelBERT vectors could model relations in a more fine-grained way than existing knowledge graphs. We focus on ConceptNet for this analysis. We cluster the RelBERT vectors of the word pairs in each relation type using HDBSCAN \cite{HDBSCAN}. We focus on three relation types: \emph{AtLocation}, \emph{CapableOf}, and \emph{IsA}. These are the relation types with the highest number of instances, among those for which HDBSCAN yielded more than one cluster. We obtained two clusters for each of these relation types. \autoref{tab:relbert:cluster} shows some examples of word pairs in each cluster. For \emph{AtLocation}, RelBERT separates the word pairs depending on whether the head denotes a living thing. On the other hand, fastText captures a surface feature and forms a cluster where the word pairs have ``zoo'' as tails. All other word pairs are mixed together in the first cluster. For \emph{CapableOf}, RelBERT\textsubscript{LARGE} again distinguishes the word pairs based on whether the head entity denotes a living thing. For RelBERT\textsubscript{BASE}, the clusters are not separated as clearly, while fastText again focuses on the presence of particular words such as ``cat'' and ``dog'' in this case. For \emph{IsA}, RelBERT\textsubscript{LARGE} yields a cluster that specifically focuses on geolocations. RelBERT\textsubscript{BASE} puts pairs with the words ``sport'' or ``game'' together. In the case of fastText, all pairs were clustered together. Overall, fastText tends to catch the surface features of the word pairs. RelBERT\textsubscript{LARGE} seems to find meaningful distinctions, although at least in these examples, they are focused on the semantic types of the entities involved rather than any specialisation of the relationship itself. The behaviour of RelBERT\textsubscript{BASE} is similar, albeit clearly noisier.

\section{Conclusion}
We have proposed a strategy for learning relation embeddings, i.e.\ vector representations of pairs of words which capture their relationship. The main idea is to fine-tune a pre-trained language model using a relational similarity dataset covering a broad range of semantic relations. In our experimental results, we found the resulting relation embeddings to be of high quality, outperforming state-of-the-art methods on most analogy questions and relation classification benchmarks, while maintaining the flexibility of an embedding model. Crucially, we found that RelBERT is capable of modelling relationships that go well beyond those that are covered by the training data, including morphological relations and relations between named entities. Being based on RoBERTa\textsubscript{LARGE}, our main RelBERT model has 354M parameters. This relatively small size makes RelBERT convenient and efficient to use in practice. Surprisingly, we found RelBERT to significantly outperform language models which are several orders of magnitude larger.

While many NLP tasks can now be solved by prompting LLMs, learning explicit representations remains important for tasks that require transparency or efficiency. For instance, we envision that RelBERT can play an important role in the context of semantic search, e.g.\ to find relevant context for retrieval augmented LMs \cite{DBLP:conf/icml/GuuLTPC20}.
Explicit representations also matter for tasks that cannot easily be described using natural language instructions, such as ontology alignment \cite{DBLP:conf/aaai/0008CA022} and completion \cite{DBLP:journals/corr/abs-2202-09791,DBLP:conf/semweb/LiBS19}, where relation embeddings should intuitively also be clearly useful. More generally, RelBERT has the potential to improve applications that currently rely on commonsense KGs such as ConceptNet, e.g.\ commonsense question answering with smaller LMs \cite{yasunaga-etal-2021-qa} and scene graph generation \cite{DBLP:conf/wacv/ChenRL23}.

\section*{Acknowledgments}
Steven Schockaert has been supported by EPSRC grants EP/V025961/1 and EP/W003309/1. Jose Camacho-Collados is supported by a UKRI Future Leaders Fellowship.

\bibliographystyle{elsarticle-num} 
\bibliography{cas-refs,anthology}

\clearpage
\appendix

\section{Model Names on HuggingFace}
\label{app:relbert:hf-name}

\begin{table}[ht]
\centering
\begin{tabular}{ll}
\toprule
Model &        Name on HuggingFace \\
\midrule
ALBERT\textsubscript{BASE}      &            \texttt{albert-base-v2} \\
BERT\textsubscript{BASE}        &            \texttt{bert-base-cased} \\
BERT\textsubscript{LARGE}       &           \texttt{bert-large-cased} \\
RoBERTa\textsubscript{BASE}     &               \texttt{roberta-base} \\
RoBERTa\textsubscript{LARGE}    &              \texttt{roberta-large} \\
GPT-2\textsubscript{SMALL}      &                       \texttt{gpt2} \\
GPT-2\textsubscript{BASE}       &                \texttt{gpt2-medium} \\
GPT-2\textsubscript{LARGE}      &                 \texttt{gpt2-large} \\
GPT-2\textsubscript{XL}       &                    \texttt{gpt2-xl} \\
GPT-J\textsubscript{125M}       &    \texttt{EleutherAI/gpt-neo-125M} \\
GPT-J\textsubscript{1.3B}       &    \texttt{EleutherAI/gpt-neo-1.3B} \\
GPT-J\textsubscript{2.7B}       &    \texttt{EleutherAI/gpt-neo-2.7B} \\
GPT-J\textsubscript{6B}         &        \texttt{EleutherAI/gpt-j-6B} \\
GPT-J\textsubscript{20B}        &    \texttt{EleutherAI/gpt-neox-20b} \\
OPT\textsubscript{125M}         &          \texttt{facebook/opt-125m} \\
OPT\textsubscript{350M}         &          \texttt{facebook/opt-350m} \\
OPT\textsubscript{1.3B}         &          \texttt{facebook/opt-1.3b} \\
OPT\textsubscript{30B}          &           \texttt{facebook/opt-30b} \\
OPT-IML\textsubscript{1.3B}     &      \texttt{facebook/opt-iml-1.3b} \\
OPT-IML\textsubscript{30B}      &   \texttt{facebook/opt-iml-30b} \\
OPT-IML\textsubscript{M-1.3B} &  \texttt{facebook/opt-iml-max-1.3b} \\
OPT-IML\textsubscript{M-30B}      &   \texttt{facebook/opt-iml-max-30b} \\
T5\textsubscript{SMALL}         &                   \texttt{t5-small} \\
T5\textsubscript{BASE}          &                    \texttt{t5-base} \\
T5\textsubscript{LARGE}         &                   \texttt{t5-large} \\
T5\textsubscript{3B}            &                      \texttt{t5-3b} \\
T5\textsubscript{11B}           &                     \texttt{t5-11b} \\
Flan-T5\textsubscript{SMALL}    &       \texttt{google/flan-t5-small} \\
Flan-T5\textsubscript{BASE}     &        \texttt{google/flan-t5-base} \\
Flan-T5\textsubscript{LARGE}    &       \texttt{google/flan-t5-large} \\
Flan-T5\textsubscript{XL}       &          \texttt{google/flan-t5-xl} \\
Flan-T5\textsubscript{XXL}      &         \texttt{google/flan-t5-xxl} \\
Flan-UL2     &            \texttt{google/flan-ul2} \\
\bottomrule
\end{tabular}
\caption{The language models used in the paper and their corresponding alias on HuggingFace model hub.}
\label{app:tab:hf-name}
\end{table}

\autoref{app:tab:hf-name} is a list of language models we used in the paper with their names on HuggingFace molde hub, where we take the pre-trained model weight.

\section{Hyperparameters of RelBERT}
\label{app:hyperparameters}

\begin{table}[ht]
\centering
\begin{tabular}{lcrrrrrr}
\toprule
Language Model              & Aggregation               & Dataset   & Loss      & Template & Epoch & Seed & Batch \\
\midrule
RoBERTa\textsubscript{BASE} & \emph{average w.o. mask}  & RelSim    & InfoNCE   & 1        & 8     & 0    & 400\\
RoBERTa\textsubscript{BASE} & \emph{average w.o. mask}  & RelSim    & InfoNCE   & 5        & 10    & 1    & 400\\
RoBERTa\textsubscript{BASE} & \emph{average w.o. mask}  & RelSim    & InfoNCE   & 5        & 9     & 2    & 400\\ \midrule
RoBERTa\textsubscript{BASE} & \emph{average w.o. mask}  & RelSim    & InfoNCE   & 1        & 10    & 0    & 25\\
RoBERTa\textsubscript{BASE} & \emph{average w.o. mask}  & RelSim    & InfoNCE   & 1        & 6     & 0    & 50\\
RoBERTa\textsubscript{BASE} & \emph{average w.o. mask}  & RelSim    & InfoNCE   & 2        & 6     & 0    & 100\\
RoBERTa\textsubscript{BASE} & \emph{average w.o. mask}  & RelSim    & InfoNCE   & 5        & 8     & 0    & 150\\
RoBERTa\textsubscript{BASE} & \emph{average w.o. mask}  & RelSim    & InfoNCE   & 1        & 8     & 0    & 200\\
RoBERTa\textsubscript{BASE} & \emph{average w.o. mask}  & RelSim    & InfoNCE   & 1        & 9     & 0    & 250\\
RoBERTa\textsubscript{BASE} & \emph{average w.o. mask}  & RelSim    & InfoNCE   & 5        & 10    & 0    & 300\\
RoBERTa\textsubscript{BASE} & \emph{average w.o. mask}  & RelSim    & InfoNCE   & 5        & 9     & 0    & 350\\ 
RoBERTa\textsubscript{BASE} & \emph{average w.o. mask}  & RelSim    & InfoNCE   & 1        & 8     & 0    & 450\\
RoBERTa\textsubscript{BASE} & \emph{average w.o. mask}  & RelSim    & InfoNCE   & 1        & 9     & 0    & 500\\
\midrule
RoBERTa\textsubscript{LARGE}& \emph{average w.o. mask}  & RelSim    & InfoNCE   & 1 & 8 & 0 & 100\\
RoBERTa\textsubscript{LARGE}& \emph{average w.o. mask}  & RelSim    & InfoNCE   & 4 & 9 & 1 & 100\\ 
RoBERTa\textsubscript{LARGE}& \emph{average w.o. mask}  & RelSim    & InfoNCE   & 4 & 8 & 2 & 100\\ \midrule
RoBERTa\textsubscript{LARGE}& \emph{average w.o. mask}  & RelSim    & InfoNCE   & 4 & 9 & 0 & 400\\ \midrule
RoBERTa\textsubscript{BASE} & \emph{average w.o. mask}  & RelSim    & InfoLOOB  &  1 & 4 & 0 & 400\\
RoBERTa\textsubscript{BASE} & \emph{average w.o. mask}  & RelSim    & Triplet   & 5 & 1 & 0 & 400\\ \midrule
RoBERTa\textsubscript{BASE} & \emph{average w.o. mask}  & NELL      & InfoNCE   & 5 & 5 & 0 & 400\\
RoBERTa\textsubscript{BASE} & \emph{average w.o. mask}  & T-REX     & InfoNCE   & 1 & 4 & 0 & 400\\
RoBERTa\textsubscript{BASE} & \emph{average w.o. mask}  & ConceptNet& InfoNCE   & 4 & 5 & 0 & 400\\ \midrule
RoBERTa\textsubscript{BASE} & \emph{average w.o. mask}  & RelSim    & InfoNCE   & 1 & 8 & 0 & 400\\
BERT\textsubscript{BASE}    & \emph{average w.o. mask}  & RelSim    & InfoNCE   & 1 & 6 & 0 & 400\\
ALBERT\textsubscript{BASE}  & \emph{average w.o. mask}  & RelSim    & InfoNCE   & 1 & 5 & 0 & 400\\
\bottomrule
\end{tabular}
\caption{The best configuration of the template and the number of epoch.}
\label{app:tab:model-config-tuned}
\end{table}

\autoref{app:tab:model-config-tuned} shows the best combination of the template and the number of epoch for each of the models used in our paper.

\end{document}